\documentclass[fleqn,10pt]{wlscirep}
\usepackage[utf8]{inputenc}
\usepackage[T1]{fontenc}
\usepackage{capt-of}
\usepackage{caption}
\usepackage{afterpage}
\usepackage{makecell}
\usepackage{multirow}
\usepackage{float}
\usepackage{enumitem}
\usepackage{fontawesome5}

\DeclareRobustCommand{\corrAuthor}{%
  \raisebox{-0.6ex}{\textsuperscript{\faEnvelope[regular]}}%
}

\definecolor{tissuelabblue}{HTML}{6352a3}
\fancypagestyle{plain}{%
  \fancyhf{}
  
  \fancyfoot[R]{\small\sffamily\fontseries{b}\selectfont\thepage/\pageref{LastPage}}
}

\title{VISTA-PATH: An interactive foundation model for pathology image segmentation and quantitative analysis in computational pathology}

\author[1]{Peixian Liang}
\author[2]{Songhao Li}
\author[1]{Shunsuke Koga}
\author[3]{Yutong Li} 
\author[1]{Zahra Alipour}
\author[4]{Yucheng Tang}
\author[4]{Daguang Xu}
\author[1,5,\corrAuthor]{Zhi Huang}
\affil[1]{Department of Pathology and Laboratory Medicine, University of Pennsylvania, PA 19104, USA}
\affil[2]{Department of Electrical and System Engineering, University of Pennsylvania, PA 19104, USA}
\affil[3]{Department of Biomedical Engineering, Georgia Institute of Technology and Emory University, Atlanta, GA 30332, USA}
\affil[4]{NVIDIA Corporation, USA}
\affil[5]{Department of Biostatistics, Epidemiology and Informatics, University of Pennsylvania, PA 19104, USA}

\affil[ ]{\corrAuthor~To whom the correspondence should be addressed: Zhi Huang (\href{zhi.huang@pennmedicine.upenn.edu}{zhi.huang@pennmedicine.upenn.edu})}

\begin{abstract}
Accurate semantic segmentation for histopathology image is crucial for quantitative tissue analysis, expert interpretation, and downstream clinical modeling. Recent segmentation foundation models have improved generalization through large-scale pretraining, yet remain poorly aligned with pathology because they either (i) lack of domain expertise in pathology; or (ii) treat segmentation as a static visual prediction task. Here we present VISTA-PATH, an interactive, class-aware pathology segmentation foundation model designed to resolve heterogeneous multi-tissue structures, incorporate expert feedback, and produce pixel-level segmentation that are directly meaningful for both quantitative measurements and clinical interpretation. VISTA-PATH jointly conditions segmentation on visual context, semantic tissue descriptions, and optional expert-provided spatial prompts, enabling precise multi-class segmentation across heterogeneous pathology images. To support this paradigm, we curate VISTA-PATH Data, a large-scale pathology segmentation corpus comprising over 1.6 million image--mask--text triplets spanning 9 organs and 93 tissue classes. Across extensive held-out and external benchmarks, VISTA-PATH consistently outperforms existing segmentation foundation models, particularly in settings with complex, overlapping tissue structures. It exceeds the second-best method by 2.7--30.9\% Dice score across 9 organs on the held-out test set, and by 3.3--16.3\% Dice across 13 organs on external datasets. Most importantly, VISTA-PATH supports dynamic human-in-the-loop refinement by propagating sparse, patch-level bounding-box annotation feedback into whole-slide, pixel-level segmentation. Across 9 evaluated datasets, VISTA-PATH improves Dice scores by 15.3\%--46.8\% via the proposed interactive refinement. Finally, we show that the high-fidelity, class-aware segmentation produced by VISTA-PATH is a preferred model for computational pathology. It improves tissue microenvironment analysis through proposed Tumor Interaction Score (TIS), which exhibits strong and significant associations with patient survival. On the TCGA-COAD cohort, VISTA-PATH improves C-index by 16.8--20.7\% over a standard multiple instance learning (MIL) model. Together, these results establish VISTA-PATH as a foundation model that elevates pathology image segmentation from a static prediction to an interactive and clinically grounded representation for digital pathology. Source code and demo can be found at \href{https://github.com/zhihuanglab/VISTA-PATH}{https://github.com/zhihuanglab/VISTA-PATH}.

\end{abstract}

\begin{document}  
\maketitle

\flushbottom
\maketitle
%
%

\section{Main}

Accurate pathology image segmentation is a crucial step for both expert interpretation and computational modeling of disease in digital pathology, and is instrumental from tumor delineation and microenvironment profiling to biomarker discovery and prognostic modeling~\cite{campanella2019clinical,coudray2018classification,echle2021deep,xu2024whole,lu2024visual,wang2024pathology}. However, existing pathology segmentation models or pipelines remain narrowly focused on specific organs, tissue types, or datasets~\cite{graham2019hover,weigert2020,zhang2023sam,chen2017deeplab}. While such specialized models can perform well in controlled settings, they often fail to generalize to new cohorts and impose a heavy annotation burden on users, who must fine-tune or retrain models for each new domain.

Recent advances in segmentation foundation models have substantially improved generalization through large-scale pretraining across diverse datasets and imaging domains~\cite{huang2023visual,xu2024whole,chen2024uni,xiang2025vision}. In the biomedical image domain, models such as MedSAM~\cite{ma2024segment} and BiomedParse~\cite{zhao2025foundation} represent important steps toward universal segmentation by reducing dataset dependency and enabling zero-shot inference. These models demonstrate that large-scale pretraining can overcome many of the limitations of dataset-specific pipelines. However, they were largely designed around radiology or natural-image tasks and therefore do not fully meet the practical demands of segmenting the heterogeneous patterns present in hematoxylin and eosin (H\&E)-stained histopathology images.

\begin{figure}[tbp]
\centering
\includegraphics[width=\textwidth,keepaspectratio]{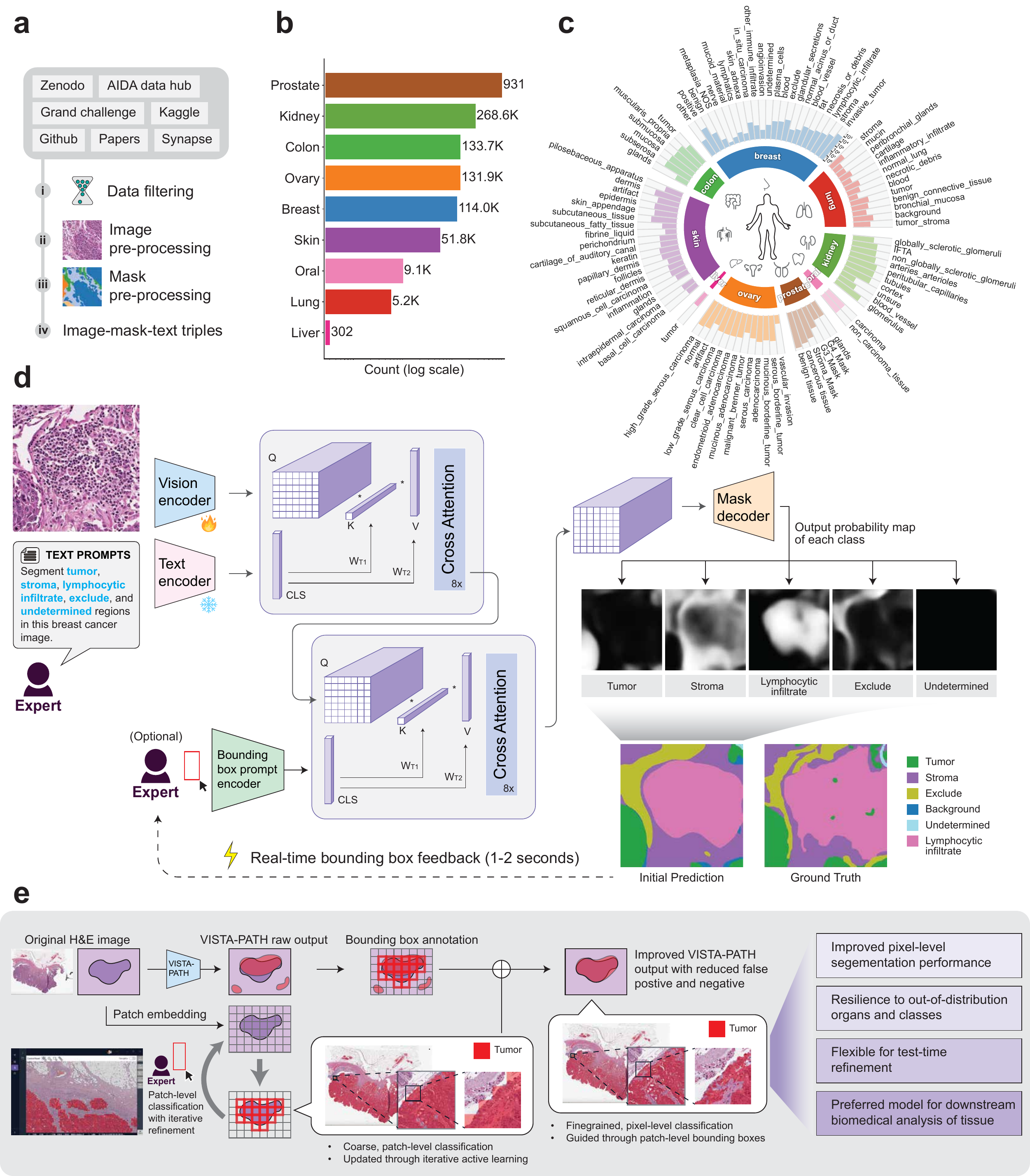}
\end{figure}


\begin{figure}[t] 
\centering
\captionof{figure}{
\textbf{Overview of the VISTA-PATH Dataset and the VISTA-PATH architecture}
\textbf{a}, Data acquisition pipeline from public pathology repositories.
\textbf{b}, Distribution of segmentation masks across organs in the VISTA-PATH Dataset (log scale).
\textbf{c}, Constructed tissue ontology organizing class labels across organs; accompanying bar plots indicate the number of images containing each tissue class (log scale).
\textbf{d}, VISTA-PATH model workflow. Given an H\&E image, class prompts, and a bounding box, vision, text, and prompt encoders extract features that are fused via cross-attention modules. A mask decoder produces class-specific segmentation maps, yielding initial predictions. Pathology experts can provide real-time bounding-box feedback to refine segmentation results.
\textbf{e}, Human-in-the-loop refinement workflow. The original H\&E image produces pixel-wise segmentations from VISTA-PATH and patch-wise segmentations via a patch embedding model. A pathologist revises annotations on a small subset of patches. These refinements are used to learn a \emph{patch-level embedding classifier}, which generalizes to infer bounding boxes across the entire image. The inferred bounding boxes are incorporated into VISTA-PATH to obtain improved pixel-level segmentation of the full image with minimal human intervention. Human-in-the-loop refinement interface is implement using TissueLab~\cite{li2025co}.
}
\label{fig:overview}
\end{figure}

Therefore, there remains a critical need to develop a highly reliable segmentation foundation model for diverse pathology images that can help facilitate many downstream analysis tasks. However, three fundamental challenges constrain the capabilities of existing segmentation models in pathology.
First, histopathology images exhibit extreme semantic and morphological heterogeneity across organs, tissue classes, staining protocols, and datasets~\cite{Lu2021CLAM,Litjens2017,liang2024enhancing,ding2025multimodal}. Models trained primarily on radiology or narrowly defined pathology datasets struggle to maintain consistent multi-class performance in such diverse settings~\cite{zhang2025generalist,zhao2025foundation}. Second, due to the high diversity of tissue microenvironment and staining \& scanning protocols, existing pathology segmentation models perform poorly on out-of-distribution data; most of these models treat segmentation as a static, one-shot prediction and provide limited mechanisms for incorporating expert feedback during or after inference~\cite{chen2025segment}. At present, pathologists routinely inspect, revise, and refine segmentation outputs based on localized visual evidence. However, existing methods rarely incorporate this expert interaction and therefore cannot improve during the test time.
Third, existing segmentation studies are often disconnected from downstream clinical tasks and lack direct evaluation of segmentation quality using clinically meaningful measures of disease aggressiveness and patient prognosis, as well as how to translate segmentation results into clinically interpretable analyses.

To address the aforementioned limitations, in this work, we introduce VISTA-PATH (\textbf{\underline{V}}isual \textbf{\underline{I}}nteractive \textbf{\underline{S}}egmentation and \textbf{\underline{T}}issue \textbf{\underline{A}}nalysis for \textbf{\underline{Path}}ology), an interactive, class-aware pathology segmentation foundation model designed to unify heterogeneous pathology image segmentation, expert-guided refinement, and clinically grounded tissue analysis within a single framework. Rather than treating segmentation as a static prediction task, VISTA-PATH elevates it to a foundational component for expert interaction and clinical discovery. The main contribution and advantage of VISTA-PATH are:
\begin{itemize}[topsep=2pt, itemsep=2pt, parsep=0pt, partopsep=0pt]
  \item \textbf{A large-scale, ontology-driven pathology segmentation dataset.}  
  We curate VISTA-PATH Data, the largest pathology segmentation corpus to date, comprising over 1.6 million image--mask--text triplets spanning 9 organs and 93 tissue classes. This resource captures both intra-organ and inter-organ morphological diversity and provides the scale and semantic coverage required to train a pathology segmentation foundation model.

  \item \textbf{Text-prompted, class-aware segmentation that supports diverse tissue concepts.}  
  VISTA-PATH introduces text-based class prompting as a first-class interface for pathology segmentation, enabling the model to segment tissue classes specified by diverse text descriptions rather than a fixed label sets. This design (i) allows flexible handling of previously unseen or hybrid class taxonomies; and (ii) allows seamless integration into large language model (LLM)-based agent systems, such as TissueLab \cite{li2025co}.

  \item \textbf{Fully interactive, real-time refinement from patch-level annotations to pixel-level segmentation.}  
  VISTA-PATH supports real-time, human-in-the-loop patch-based active learning, in which sparse expert corrections at the patch level are propagated to whole-slide, pixel-level predictions. By converting localized annotations into class-aware spatial prompts, the framework enables efficient global refinement without dense relabeling or model retraining, and can greatly improve the segmentation performances.

  \item \textbf{High-fidelity segmentation supports clinical discovery.}
VISTA-PATH extends segmentation beyond accuracy-oriented evaluation by making it clinically operationalizable. With high-fidelity, class-aware tissue maps, VISTA-PATH can extract better and interpretable morphological features that helps quantify spatial tissue organization and tumor–microenvironment interactions, supporting downstream analyses that directly link tissue morphology to patient outcomes.

\end{itemize}

Across extensive held-out and external evaluations, VISTA-PATH demonstrates robust and consistent improvements over existing segmentation foundation models, particularly in challenging settings with complex and overlapping tissue structures. The resulting high-fidelity, class-aware tissue maps enable clinically meaningful downstream analyses. In particular, Tumor Interaction Score (TIS) derived from VISTA-PATH segmentation shows strong and significant associations with colon cancer patient survival, improving the C-index by 16.8--20.7\% over a standard multiple instance learning (MIL) model on the TCGA-COAD cohort. Fully integrated into TissueLab pathology AI platform \cite{li2025co}, VISTA-PATH is readily accessible for various research applications. Source code and an interactive demo are also available at \href{https://github.com/zhihuanglab/VISTA-PATH}{https://github.com/zhihuanglab/VISTA-PATH}.

\section{Results}
\subsection{Curating a large image--mask--text segmentation dataset across 9 organs}

Although deep learning has significantly advanced medical image analysis, large-scale and high-quality pathology image segmentation datasets remain notably scarce. Compared to other domains such as natural scene or radiology, the pathology faces unique challenges: whole-slide images are at ultra-high resolution, tissue structures are morphologically diverse, and semantic definitions of cellular and tissue entities often vary across datasets. Existing resources are typically fragmented, small in scale, and lack consistent pixel-level annotations, making it difficult to train and evaluate pathology segmentation foundation models. This scarcity of well-curated pathology datasets represents a critical bottleneck for progress quantitative analysis in computational pathology.

To address this limitation and support scalable pathology segmentation across diverse organs and tissue types, we constructed VISTA-PATH Data, a large-scale pathology image segmentation corpus constructed from 22 publicly available datasets (\textbf{Fig.~\ref{fig:overview}a}) from different sources across Zenodo, AIDA data hub, Competition and Challenge (\textit{e.g.}, Kaggle), GitHub, publications, etc. Unlike conventional datasets that are restricted to a single organ or narrowly defined tissue class, VISTA-PATH Data aggregates heterogeneous pathology images spanning organs, tissue types, and image types. Comprising 1{,}645{,}706 image--mask--class triplets, VISTA-PATH Data is by far the largest and most diverse segmentation dataset in routine H\&E pathology domain with both tissue-level and instance-level segmentation annotations. 

In VISTA-PATH Data, each pathology image can contain multiple tissue classes and multiple instances per class. Across the entire corpus, the dataset covers 9 major organs and 93 tissue classes and including tumor, normal anatomical structures, and microenvironment-related components (\textbf{Fig.~\ref{fig:overview}b--c}). This diversity enables computational modeling of both intra-organ and inter-organ morphological variation. To ensure high data quality, we applied a rigorous filtering process that retained only datasets with precise pixel-wise annotations and high-quality images. Specifically, we excluded low-intensity images and coarse annotations (for example, those generated using convex-hull labeling). We then applied a unified preprocessing pipeline across all retained datasets to ensure consistency and comparability (See \textbf{Methods} section for details). The resulting dataset provides both the scale and semantic diversity required to train a pathological segmentation foundation model.

After cleaning, preprocessing, and organization, we randomly sampled 95\% pathology images from each source and mix them together to form the training data (see \textbf{Fig.~\ref{fig:overview}a--c} and \textbf{Extended Data Fig.~\ref{sup:fig1}}) comprises 1,645,706 patch-level samples spanning 93 classes. The rest of the datasets are held out as test set for in-domain evaluation purposes with a total of 77,107 samples from 69 classes. Test images are drawn from the same datasets used for training but consist of non-overlapping images, enabling a controlled evaluation of segmentation accuracy under in-distribution conditions. Training and test sets are split at the original image level to ensure that no derived patch samples from the same image appear in both splits. Data sources of VISTA-PATH Data can be found in \textbf{Extended Data Table~\ref{sup:internal_data_source}}, and detailed statistics are presented in \textbf{Extended Data Fig.~\ref{sup:fig2}}.

To assess robustness of VISTA-PATH under domain shift, we further collect external datasets that come from completely different sources from VISTA-PATH data. Details can be found in \textbf{Extended Data Table~\ref{sup:external_data_source}} and \textbf{Extended Data Fig.~\ref{sup:fig3}}. In summary, external datasets include 66,355 samples, 82 tissue classes, and 13 organs from three fundamentally different annotation regimes: additional H\&E pathology segmentation datasets, Visium HD spatial transcriptomics-guided H\&E segmentation datasets, and Xenium single-cell spatial transcriptomics-guided H\&E segmentation datasets. For Visium HD data, spatial bins were clustered based on gene expression profiles and manually assigned semantic tissue labels by pathologists; bin-level annotations were subsequently converted into dense pixel-wise segmentation maps using platform-provided spatial geometry. For Xenium data, single-cell molecular profiles and spatial coordinates were jointly clustered, and cluster labels were propagated to pixel-level maps using platform-provided cell segmentation contours (see \textbf{Methods} section for details). This setting provide diverse and comprehensive analysis dimensions to probe VISTA-PATH’s capability on generalization across organs, tissue classes, and imaging paradigms.

\subsection{Developing VISTA-PATH foundation model for pathology segmentation}
Based on the large 1,645,706 training data, we innovated and trained VISTA-PATH, an interactive, class-aware, semantic segmentation pathology foundation model that unifies morpho-textural evidence, semantic tissue identity, and spatial context within a unified architecture (\textbf{Fig.~\ref{fig:overview}d}). Rather than treating class recognition and spatial localization as separate or sequential processes, VISTA-PATH conditions segmentation simultaneously on what tissue is present and where it is located, enabling robust multi-class reasoning in heterogeneous pathology images. To achieve this, VISTA-PATH comprises four components: \textbf{a trainable image encoder}, \textbf{a frozen text encoder}, \textbf{an optional bounding-box prompt encoder}, and \textbf{a mask decoder}. The image and text encoders are initialized from the pretrained PLIP~\cite{huang2023visual} vision--language model, providing aligned visual--semantic representations specialized for histopathology. Semantic tissue identity is specified through class-aware text prompts, which encode the biological meaning of each target tissue class. Optionally, bounding-box prompts are provided as spatial priors and encoded by a SAM-based prompt encoder~\cite{kirillov2023segment}. These visual, textual, and spatial representations are fused through cross-attention, allowing the model to resolve tissue identity and spatial extent within a unified latent space (\textbf{Fig.~\ref{fig:overview}d}).
Specifically, given an input image and a target tissue description, VISTA-PATH produces a class-conditioned, pixel-wise segmentation mask. When a bounding-box prompt is available, it provides an additional supervision. This joint conditioning on semantic class and bounding-box guidance enables VISTA-PATH to accurately segment complex tissue microenvironments.

VISTA-PATH model is trained under a class-conditioned segmentation formulation, in which each forward pass predicts the foreground--background mask for a specified tissue class. To prevent over-reliance on spatial priors and promote robustness, bounding-box prompts are generated by ground truth masks and randomly dropped during training (See \textbf{Methods} for details). At inference time, multi-class segmentation is obtained by aggregating the class-wise predictions into a coherent tissue map.

Since the model is trained using random bounding-box prompts, VISTA-PATH naturally supports human-in-the-loop refinement at test time to further rapidly reduce false positives and false negatives. Implemented within the TissueLab platform \cite{li2025co}, this iterative interaction enables performance improvements with minimal annotation effort (\textbf{Fig.~\ref{fig:overview}e}) and facilitates adaptation of the foundation model to out-of-distribution cases. Given a raw H\&E whole-slide image, VISTA-PATH first generates pixel-wise segmentation predictions. In parallel, a complementary patch-based embedding model (MUSK) \cite{xiang2025vision} produces patch-level vector representations that are used to train a lightweight patch classification model. A pathologist can then add or revise annotations on a small subset of informative patches, which are used to train the classifier. The trained classifier is rapidly applied across the whole-slide image to produce updated patch-level segmentation results. Because training occurs only after patch embeddings are computed, this process can be performed in real time. After each update, VISTA-PATH converts the refined patch-level masks into bounding-box prompts, which are fed back into the segmentation model to further improve pixel-level predictions, resulting in progressively refined whole-slide segmentation. The model was optimized using the AdamW optimizer with a learning rate of $5\times10^{-5}$ for 10 epochs under mixed-precision training on NVIDIA H200 GPUs.

\subsection{VISTA-PATH enables joint semantic--spatial multi-class segmentation in heterogeneous pathology}

After the model trained, we evaluate whether VISTA-PATH fulfills the central requirement of a pathology segmentation foundation model: robust joint semantic--spatial multi-class segmentation under extreme tissue heterogeneity. To this end, we compare VISTA-PATH against two representative segmentation foundation models: 
BiomedParse, which conditions segmentation on class-aware text without explicit spatial priors, and MedSAM, which uses spatial prompts without semantic competition between tissue classes (\textbf{Fig.~\ref{fig2:internal_results}a}). BiomedParse conditions segmentation solely on class-aware textual prompts paired with the input image. While this design provides semantic flexibility, the absence of explicit spatial grounding (\textit{e.g.}, bounding-box annotation) forces the model to infer tissue localization from global visual cues, which is unreliable in histopathology images where visually similar patterns recur across different anatomical contexts. MedSAM (\textbf{Fig.~\ref{fig2:internal_results}b}), by contrast, incorporates bounding-box prompts that provide stronger spatial guidance but produces binary, class-isolated masks. However, when multiple tissue types coexist within overlapping spatial regions which is a common scenario in pathology, the model lacks a mechanism to resolve class-level ambiguity, leading to misclassification or fragmented predictions. VISTA-PATH (\textbf{Fig.~\ref{fig2:internal_results}c}) resolves both limitations by jointly conditioning segmentation on class-aware text prompts and corresponding spatial prompts within a unified vision--language architecture, enabling the model to reason simultaneously about what tissue is present and where it is located. PathSegmenter~\cite{chen2025segment} was not included in our comparison, as the pathology-specific pretrained models and code needed for evaluation were not publicly available at the time of our experiments.

We first evaluate VISTA-PATH on held-out test sets from VISTA-PATH Data to assess in-distribution multi-class segmentation under controlled but heterogeneous conditions. Across the held-out internal datasets, VISTA-PATH achieves the highest average Dice scores with reduced variance and tighter 95\% confidence intervals than both MedSAM and BiomedParse (\textbf{Fig.~\ref{fig2:internal_results}d}). VISTA-PATH outperforms MedSAM on the majority of datasets, with average gains ranging from 0.006--0.414 Dice. Compared with BiomedParse, VISTA-PATH achieves higher Dice scores on all but one dataset (GlaS), with average gains ranging from 0.010--0.668 Dice. These gains are observed across datasets spanning more than two orders of magnitude in size, from small cohorts to collections containing tens of thousands of images, demonstrating that the joint semantic--spatial formulation scales without loss of fidelity.

To demonstrate the advantage of VISTA-PATH under a fair end-to-end comparison, we additionally trained an organ-specific segmentation model, Res2Net~\cite{gao2019res2net}, on each organ in VISTA-PATH Data and compared its performance with VISTA-PATH without bounding-box prompts (VISTA-PATH w/o box). As results provided in \textbf{Extended Data Table~\ref{sup:table_internal}}, although Res2Net is trained separately for each dataset under optimal, task-specific conditions, VISTA-PATH achieves a substantially higher average Dice (0.772 for VISTA-PATH versus 0.521 for Res2Net, $P < 1 \times 10^{-3}$) and outperforms Res2Net on 19 of the 21 testing datasets, indicating that joint semantic--spatial pretraining can surpass even dataset-specific fully fine-tuned model. Moreover, while removing bounding-box prompts degrades some performances (0.698 for VISTA-PATH w/o box versus 0.772 for VISTA-PATH, $P < 1 \times 10^{-3}$), VISTA-PATH w/o box still remarkably outperforms both MedSAM and BiomedParse on average (0.698 for VISTA-PATH w/o box versus 0.581 for MedSAM versus 0.379 for BiomedParse), demonstrating that VISTA-PATH retains strong generalization even without additional, explicit bounding-box guidance.

\begin{figure}[tbp]
\centering
\includegraphics[width=\textwidth,height=\textheight,keepaspectratio]{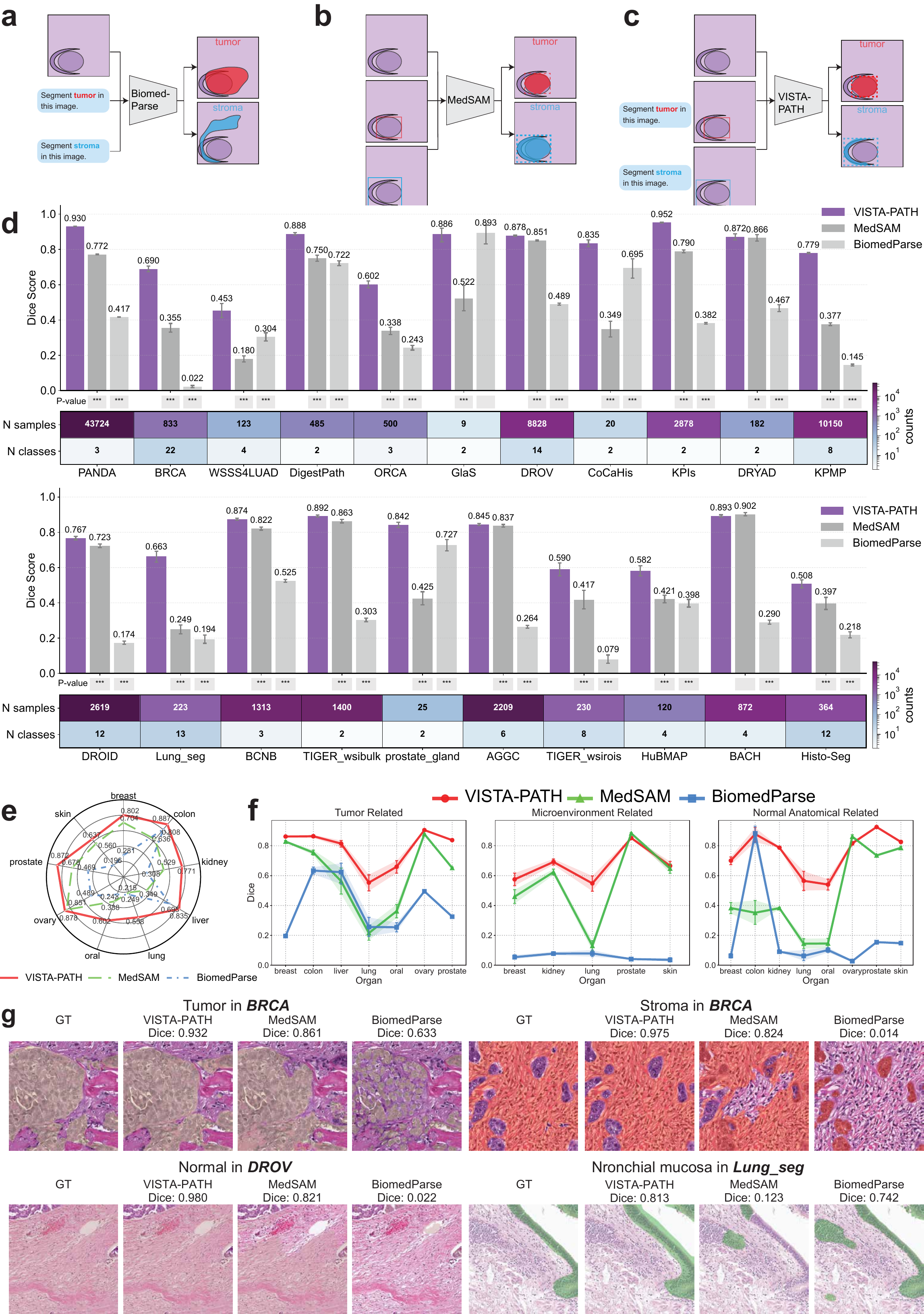}
\end{figure}

\begin{figure}[t] 
\centering
\captionof{figure}{
\textbf{Comparison of VISTA-PATH with existing segmentation foundation models.}
\textbf{a--c}, Architectural comparison of representative segmentation foundation models.
BiomedParse (a) accepts an image and class-aware textual prompts, but lacks explicit spatial priors (\textit{e.g.}, bounding boxes).
MedSAM (b) takes an image and class-specific bounding boxes as input to output binary segmentations; however, this design is prone to errors when multiple tissue types share overlapped bounding boxes.
In contrast, VISTA-PATH (c) jointly leverages class-aware text prompts and corresponding bounding boxes to enable spatially informed, precise \emph{multi-class} segmentation.
\textbf{d--f}, Quantitative segmentation performance on the VISTA-PATH held-out test set.
(d) Bar plots comparing performance across individual datasets. For each dataset, the number of images, number of classes, error bars (95\% confidence intervals), and $P$ values are reported. Statistical significance is assessed using two-sided Student's $t$-test ($^{*}P < 0.05$; $^{**}P < 1 \times 10^{-2}$; $^{***}P < 1 \times 10^{-3}$).
(e) Radar plot comparing model performance across $9$ organ types.
(f) Line plots showing Dice scores with 95\% confidence intervals across different organs, grouped by tumor-related, normal-related, and microenvironment-related tissue categories.
\textbf{g}, Case studies illustrating representative segmentation results.
}
\label{fig2:internal_results}
\end{figure}

When evaluating performance at the organ level, VISTA-PATH's advantage holds consistently across all nine organs (\textbf{Fig.~\ref{fig2:internal_results}e} and \textbf{Extended Data Table~\ref{sup:organ_internal}}). Specifically, VISTA-PATH improves over second-best model MedSAM by 0.027--0.486 Dice, with particularly large gains on heterogeneous tissues such as liver (0.835 for VISTA-PATH versus 0.349 for MedSAM, $P < 1 \times 10^{-3}$), lung (0.558 for VISTA-PATH versus 0.215 for MedSAM, $P < 1 \times 10^{-3}$), oral (0.602 for VISTA-PATH versus 0.338 for MedSAM, $P < 1 \times 10^{-3}$), and kidney (0.771 for VISTA-PATH versus 0.529 for MedSAM, $P < 1 \times 10^{-3}$). In contrast, MedSAM performs comparably only on relatively homogeneous organs such as ovary (0.851 for MedSAM versus 0.878 for VISTA-PATH) but degrades sharply on more heterogeneous tissues such as oral (0.338 for MedSAM versus 0.602 for VISTA-PATH) and lung (0.215 for MedSAM versus 0.558 for VISTA-PATH), consistent with the limitations of its binary, class-isolated prediction paradigm. In contrast, BiomedParse exhibits degradation performances across organs, with VISTA-PATH exceeding it by 0.080--0.521 Dice, including dramatic gaps on kidney (0.771 for VISTA-PATH versus 0.308 for BiomedParse, $P < 1 \times 10^{-3}$), breast (0.802 for VISTA-PATH versus 0.281 for BiomedParse, $P < 1 \times 10^{-3}$), prostate (0.872 for VISTA-PATH versus 0.469 for BiomedParse, $P < 1 \times 10^{-3}$), and skin (0.637 for VISTA-PATH versus 0.196 for BiomedParse, $P < 1 \times 10^{-3}$). Only a small subset of datasets such as colon (0.808 for BiomedParse versus 0.887 for VISTA-PATH) and liver (0.695 for BiomedParse versus 0.835 for VISTA-PATH) show comparable results, reflecting BiomedParse's limited segmentation ability in pathology and its pretraining bias toward a few of pathology tissue types.

We further stratify the results by tissue category, including tumor-related, normal anatomical, and microenvironment components, and observe that VISTA-PATH continues to exhibit robust advantages beyond organ-level comparisons (\textbf{Fig.~\ref{fig2:internal_results}f} and \textbf{Extended Data Table~
\ref{sup:organ_tissue_internal}}). Across tumor-related, microenvironment-related, and normal anatomical tissues, VISTA-PATH consistently achieves the highest Dice scores for the vast majority of organs, outperforming both MedSAM and BiomedParse. While occasional near-parity or reversals are observed for MedSAM on prostate microenvironment and ovary normal anatomy, these represent isolated cases. Overall, the results demonstrate that VISTA-PATH's advantage extends beyond specific organs to a broad spectrum of tissue types, highlighting its robustness for multi-class, multi-tissue pathology segmentation.

Representative examples in \textbf{Fig.~\ref{fig2:internal_results}g} further illustrate the advantage of VISTA-PATH. BiomedParse frequently produces spatially diffuse or incomplete segmentations, because semantic prompts alone cannot resolve local spatial ambiguity. MedSAM, despite strong localization, often misclassifies tissues when bounding boxes include heterogeneous tissue content. In contrast, VISTA-PATH preserves both spatial precision and semantic separation, enabling precise segmentation in these challenging regions.

\begin{figure}[tbp]
\centering
\includegraphics[width=\textwidth]{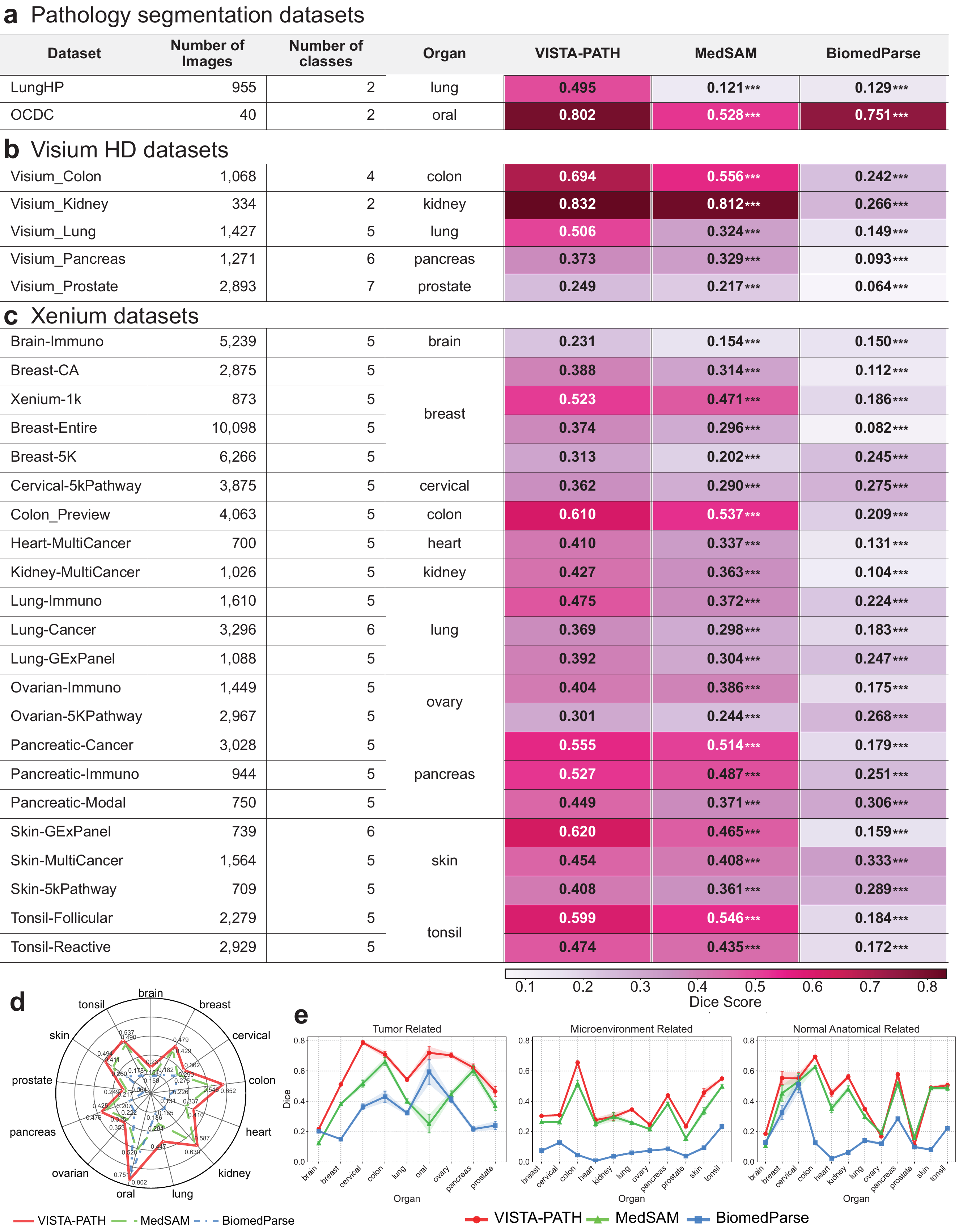}
\captionof{figure}{
\textbf{VISTA-PATH generalizes across annotation protocols, organs, and tissue types on external datasets.}
\textbf{a}, Segmentation evaluation of LungHP and OCDC datasets. 
\textbf{b}, Segmentation evaluation of 5 Visium HD datasets. 
\textbf{c}, Segmentation evaluation of 22 Xenium datasets.
\textbf{d}, Radar plot comparing model performance across 13 organ types.
\textbf{e}, Line plots showing Dice scores with 95\% confidence intervals across different organs, grouped by tumor-related, microenvironment-related, and normal anatomical tissue categories. 
}
\label{fig3:external_results}
\end{figure}

\subsection{VISTA-PATH generalizes across organs, tissue classes, and annotation paradigms}

To evaluate robustness under domain shift, we evaluate VISTA-PATH on an external benchmark comprising 66,355 images, 82 tissue classes, and 13 organs from three fundamentally different annotation regimes: independent pathology segmentation datasets, Visium HD spatial transcriptomics-guided H\&E segmentation datasets, and Xenium single-cell spatial transcriptomics-guided H\&E segmentation datasets (\textbf{Fig.~\ref{fig3:external_results}}, \textbf{Extended Data Fig.~\ref{sup:fig3}}). These datasets probe not only new organs and tissue types, but also distinct ways of defining tissue labels, ranging from expert-annotated histology to gene-expression--derived spatial domains.

Across external pathology datasets (\textbf{Fig.~\ref{fig3:external_results}a}), VISTA-PATH substantially outperforms MedSAM and BiomedParse. On LungHP\cite{li2020deep} (Lung histopathology dataset), VISTA-PATH achieves a Dice score of 0.495, more than quadrupling MedSAM (Dice=0.121) and BiomedParse (Dice=0.078). On OCDC~\cite{dos2023hematoxylin} (oral cavity-derived cancer whole-slide images dataset), VISTA-PATH reaches 0.802, compared to 0.528 and 0.203, respectively, demonstrating strong generalization even in low-sample, low-class-count settings.

The advantage of VISTA-PATH becomes more significant on spatial transcriptomics-guided H\&E segmentation datasets (\textbf{Fig.~\ref{fig3:external_results}b}) when introducing more complex, multi-class tissue structure. 
In our experiment, VISTA-PATH achieves Dice scores of 0.832 on Visium\_Kidney and 0.694 on Visium\_Colon, exceeding MedSAM by clear margins and outperforming BiomedParse by more than 0.5 Dice in several cases. Performance gaps widen as the number of tissue classes increases, indicating that joint semantic--spatial conditioning becomes increasingly critical under heterogeneity. On twenty-two Xenium datasets (\textbf{Fig.~\ref{fig3:external_results}c}), which provide dense, single-cell--derived multi-class labels, VISTA-PATH again delivers consistently superior performance across breast, lung, colon, pancreas, skin, tonsil, and ovary. Dice scores frequently exceed 0.5 in these challenging settings, whereas MedSAM and BiomedParse degrade sharply, particularly when six or more tissue classes coexist. We further evaluated VISTA-PATH without bounding-box prompts (VISTA-PATH w/o box) on the external benchmarks (\textbf{Extended Data Table~\ref{sup:table_external}}). Although performance is reduced without spatial guidance, the model still achieves an average Dice of 0.340, remaining competitive with MedSAM (Dice=0.373) and clearly outperforming BiomedParse (Dice=0.199).

Aggregated analyses across external datasets (\textbf{Fig.~\ref{fig3:external_results}d--g} and \textbf{Extended Data Tables~\ref{sup:organ_external}--\ref{sup:organ_tissue_external}}) show that VISTA-PATH consistently attains the highest Dice scores across the 13 evaluated organs and for most tumor-related, microenvironment-related, and normal anatomical tissues. In particular, VISTA-PATH maintains clear advantages over both MedSAM and BiomedParse on heterogeneous organs such as lung (0.447 for VISTA-PATH versus 0.284 for MedSAM versus 0.186 for BiomedParse), colon (0.652 for VISTA-PATH versus 0.546 for MedSAM versus 0.226 for BiomedParse), kidney (0.630 for VISTA-PATH versus 0.587 for MedSAM versus 0.185 for BiomedParse), and skin (0.494 for VISTA-PATH versus 0.411 for MedSAM versus 0.260 for BiomedParse), and this superiority extends across most tumor-related and microenvironment-related tissue compartments (\textbf{Extended Data Table~\ref{sup:organ_tissue_external}}). Although isolated near-parity cases occur for MedSAM in a small number of normal anatomical tissues (for example, ovary and prostate), the overall pattern remains strongly in favor of VISTA-PATH. These results indicate that coupling class-aware semantic conditioning with explicit bounding-box annotations yields stable and accurate multi-class segmentation across highly heterogeneous external pathology and spatial-omics datasets.

\subsection{Real-time human-in-the-loop bounding-box prompts enables efficient whole-slide refinement}

In clinical practice, segmentation errors are inevitable, and pathologists routinely inspect and correct automated outputs during analysis. An effective pathology foundation model must therefore support interactive expert-guided refinement, enabling localized human input to be efficiently propagated to global, whole-slide predictions. We evaluate whether VISTA-PATH can serve as such an interactive segmentation substrate using a real-time human-in-the-loop active learning-based whole-slide refinement pipeline using our in-house TissueLab platform \cite{li2025co} (\textbf{Fig.~\ref{fig4:human-in-the-loop}a}).

Starting from a whole-slide image, a patch-level classifier generates an initial coarse segmentation by extracting patch embeddings and predicting tissue classes. Pathologists then review a small subset of patches and provide quick corrections through active learning. These refined annotations are used to update the patch-level classifier, yielding improved patch-wise predictions. This iterative process is repeated for several rounds until visually satisfactory convergence is achieved. Then, the refined patch-level segmentation masks are converted into class-aware bounding boxes, which, together with the raw image and semantic tissue descriptions, are used as bounding-box prompts for VISTA-PATH. In this way, sparse expert feedback is transformed into spatially and semantically grounded prompts that drive pixel-level refinement across the entire whole-slide image.

We select spatial transcriptomics-derived segmentation datasets, including Visium HD and Xenium, as evaluation datasets because they are unseen during training and provide accurate pixel-level definitions. \textbf{Fig.~\ref{fig4:human-in-the-loop}b--c} and \textbf{Extended Data Tables~\ref{tab:dice_visium}--\ref{tab:dice_xenium}} report the human-in-the-loop performance on Visium HD and Xenium datasets. Across all datasets, iterative refinement rapidly increases patch-level accuracy within a small number of interaction rounds, typically raising Dice from initial values around 0.1--0.4 to 0.6--0.8 within 4--5 rounds. Importantly, VISTA-PATH consistently translates these patch-level improvements into superior pixel-wise segmentations, maintaining a clear margin over both the patch-level segmentation and MedSAM at nearly every interaction stage. For example, on Visium\_Colon, VISTA-PATH increases from 0.629 Dice at initialization to 0.867 Dice after iterative refinement, whereas the patch-level classifier and MedSAM reach only 0.808 Dice and 0.769 Dice, respectively. Similar trends are observed on Visium\_Prostate, where VISTA-PATH rises from 0.424 to 0.810, compared with 0.763 for the patch classifier and 0.775 for MedSAM, as well as on Visium\_Kidney, where VISTA-PATH improves from 0.118 to 0.563, exceeding both the patch-level baseline (Dice=0.534) and MedSAM (Dice=0.462). Compared to VISTA-PATH and patch-level segmentation results, MedSAM shows unstable behavior: despite receiving the same refined bounding boxes, its pixel-level outputs are often comparable to or worse than its patch-level predictions, reflecting its inability to translate class-conditioned spatial feedback into accurate segmentation results.

On the Xenium datasets, where dense, single-cell--derived annotations introduce substantial class and boundary complexity, our conclusions also hold. On Lung-Immuno, VISTA-PATH reaches a Dice of 0.745, compared with 0.716 for the patch classifier and 0.626 for MedSAM, and on Breast-5K, VISTA-PATH achieves 0.693, exceeding both the patch baseline (Dice=0.610) and MedSAM (Dice=0.578), respectively. In contrast, MedSAM often fails to fully exploit the refined bounding boxes, producing pixel-level segmentations that perform worse than both VISTA-PATH and, in some cases, even the patch-level predictions.

Notably, performances usually saturate after 4--5 iterations using only about 1,000 annotated patches, highlighting the efficiency of the proposed interactive, human-in-the-loop active learning paradigm. Representative examples in \textbf{Fig.~\ref{fig4:human-in-the-loop}d--e} further illustrate that VISTA-PATH produces accurate, coherent pixel-level segmentations for fine-grained and complex tissue structures. These results demonstrate that VISTA-PATH enables expert feedback to be operationalized at whole-slide scale, closing the loop between human expertise and foundation-model segmentation.

\subsection{VISTA-PATH produces accurate biomarker for enhanced outcome prediction}

In oncology, pathology remains the gold standard for disease diagnosis, assessment of progression, and treatment planning~\cite{dang2025deep}. These clinical decisions are grounded in the interpretation of tissue morphology and spatial organization, skills that pathologists develop through extensive training and experience. While recent segmentation foundation models have improved the accuracy of pixel-level tissue delineation, their outputs are often treated as intermediate visual predictions rather than as quantitative substrates for downstream computational analysis. As a result, improvements in segmentation accuracy do not automatically translate into measurable gains in clinical or biological insight, creating a disconnect between algorithmic performance and practical utility in digital pathology.

\begin{figure}[H]
\centering
\includegraphics[width=\textwidth]{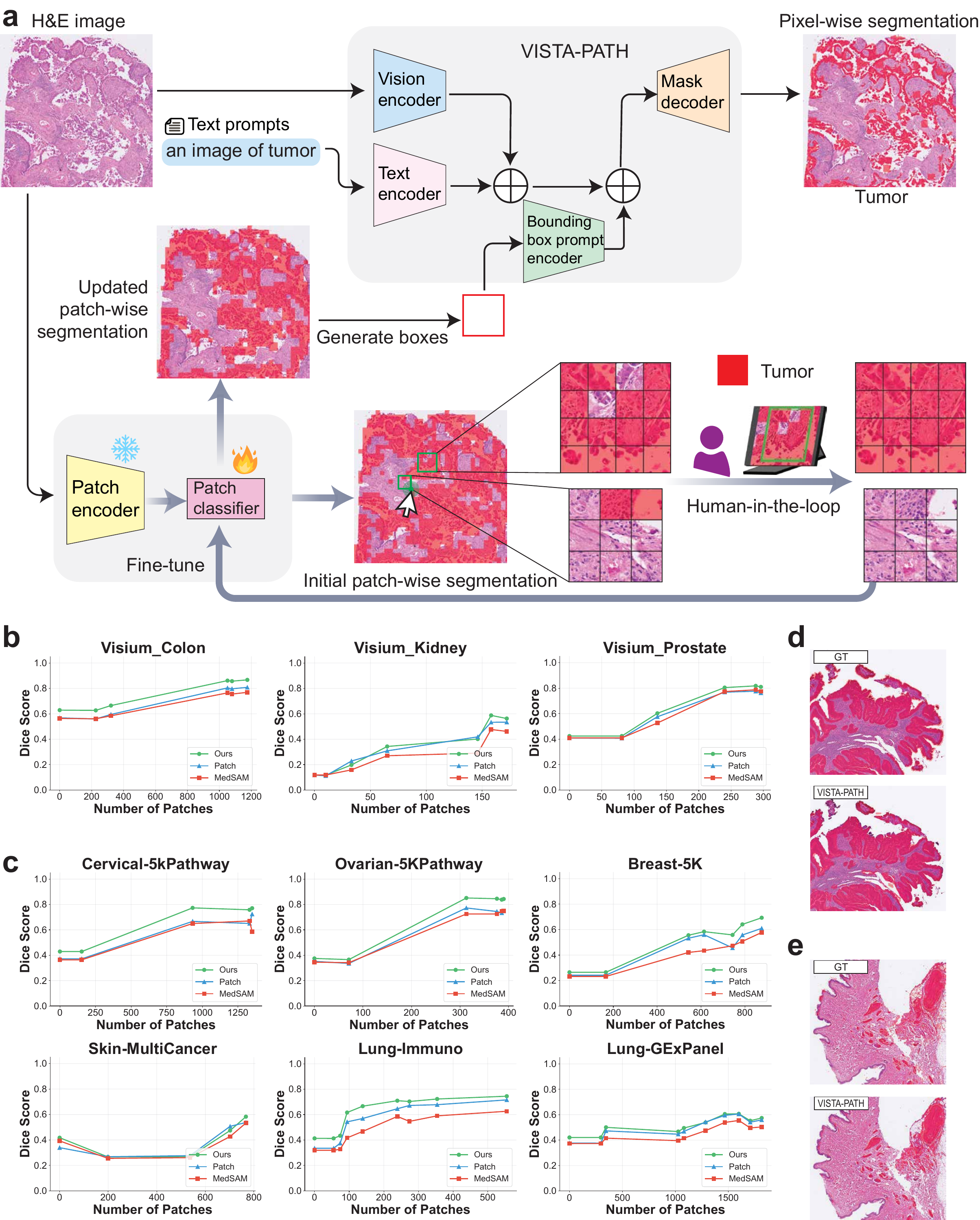}
\end{figure}

\clearpage

\begin{figure}[t] 
\centering
\captionof{figure}{
\textbf{Human-in-the-loop enhances VISTA-PATH segmentation performance.}
\textbf{a}, Starting from a H\&E image, a patch-level segmentation model produces an initial segmentation. A pathologist then reviews these predictions and provides corrective annotations for selected patches. These corrected annotations are used to fine-tune the patch-level classifier, yielding updated patch-level segmentation results. The updated patch-level segmentations are subsequently converted into bounding-box prompts and provided to VISTA-PATH to generate the final pixel-wise segmentation.
\textbf{b}, Pixel-wise segmentation performance improves on the Visium HD dataset.
\textbf{c}, Pixel-wise segmentation performance improves on the Xenium dataset.
\textbf{d}, Representative visual examples showing our pixel-wise segmentation performance on the Visium\_Colon dataset.
\textbf{e}, Representative visual examples showing our pixel-wise segmentation performance on the Skin-MultiCancer dataset. GT: Ground truth.
}
\label{fig4:human-in-the-loop}
\end{figure}

In this section, we explore whether the class-aware, high-fidelity segmentation produced by VISTA-PATH can be operationalized into clinically meaningful morphological measurements and served as a preferred tool for future research. To this end, we introduce the \textbf{Tumor Interaction Score (TIS)}, an interpretable biomarker that quantifies tumor coherence versus infiltration using both pixel-level and patch-level tumor predictions. TIS translates the spatial organization of tumor tissue into a single quantitative index:
\begin{equation}
\mathrm{TIS} = \frac{\sum_{i=1}^{N} \left| P_i \cap S \right|}{\sum_{i=1}^{N} \left| P_i \right|},
\end{equation}
where $P_i$ denotes the $i$-th patch identified as tumor by the patch-level classifier, $S$ represents the pixel-wise tumor segmentation mask derived from VISTA-PATH, and $\lvert \cdot \rvert$ denotes area measured in pixels. TIS is an explainable, clinically interpretable index: high TIS reflects cohesive and spatially confined tumor architecture, whereas low TIS may capture either (i) boundary fragmentation and invasive growth, which are morphological signatures of aggressive tumors and adverse prognosis; or (ii) a dynamic tumor–immune microenvironment (\textit{e.g.}, lymphocyte infiltration), which is commonly associated with active immune surveillance and distinct prognostic implications depending on tumor context.

To evaluate the clinical value of this segmentation-derived biomarker, we incorporate TIS into a survival prediction pipeline (\textbf{Fig.~\ref{fig5:TCGA-COAD}a}). Raw whole-slide images are first processed by a patch-level encoder--classifier to identify candidate tumor regions, which are then provided as bounding-box prompts to VISTA-PATH to generate high-fidelity pixel-wise tumor maps. TIS is computed from these outputs and used as input to a multi layer perceptron (MLP)~\cite{cybenko1989approximation} to predict patient-level survival risk (Cox proportional hazards regression). We evaluate this framework on the TCGA-COAD cohort~\cite{weinstein2013cancer} (\textbf{Fig.~\ref{fig5:TCGA-COAD}b--d}), using patient-level splits across the TCGA-AZ (or AZ) and TCGA-A6 (or A6) sites, and compare against two baselines: (i) a standard ABMIL~\cite{ilse2018attention} pipeline using the same MUSK~\cite{xiang2025vision} embeddings, and (ii) an otherwise identical pipeline in which VISTA-PATH is replaced by MedSAM. For this experiment, BiomedParse is not comparable because it does not support bounding-box--guided segmentation.

As shown in \textbf{Fig.~\ref{fig5:TCGA-COAD}c--d}, TIS substantially enhances survival prediction performance. Our VISTA-PATH-based pipeline achieved superior concordance indices~\cite{harrell1996multivariable} and log-rank t-test $P$-values (C-index = 0.678 in A6 and 0.739 in AZ; $P$ = $2.7 \times 10^{-3}$), outperforming both ABMIL (C-index = 0.510 in A6 and 0.533 in AZ; $P$ = $0.908$) and the MedSAM-based variant (C-index = 0.621 in A6 and 0.696 in AZ; $P$ = $0.399$). Ablation studies (\textbf{Fig.~\ref{fig5:TCGA-COAD}e}) further underscore the robustness of our approach: when the MLP is removed, the performance of the MedSAM variant degrades substantially (with the $P$-value increasing from 0.399 to 0.753), whereas the VISTA-PATH-derived TIS retains strong predictive power ($P$ = $3.18\times 10^{-2}$). These results demonstrate that class-aware segmentation from VISTA-PATH can be directly transformed into robust, interpretable, and clinically meaningful biomarkers, elevating segmentation from a static prediction to a quantitative instrument for clinical inference. \textbf{Fig.~\ref{fig5:TCGA-COAD}f} shows a strong inverse association between the Tumor Interaction Score (TIS) and model-predicted patient risk. \textbf{Extended Data Fig.~\ref{sup:fig4}} presents three representative examples spanning low, intermediate, and high TIS values derived from VISTA-PATH. Low TIS (TIS=0.607) reflects infiltrative tumor budding and elevated risk; intermediate TIS (TIS=0.797) corresponds to spatially confined tumors with the lowest predicted risk; and very high TIS (TIS=0.857) captures large, cohesive tumor masses in which excessive tumor burden leads to an increase in predicted risk. These findings indicate that TIS functions as an interpretable morphological biomarker that integrates tumor boundary coherence and tumor burden, enabling clinically meaningful survival prediction that outperforms other methods and can be accurately derived only from VISTA-PATH.

\begin{figure}[H]
\centering
\includegraphics[width=\textwidth]{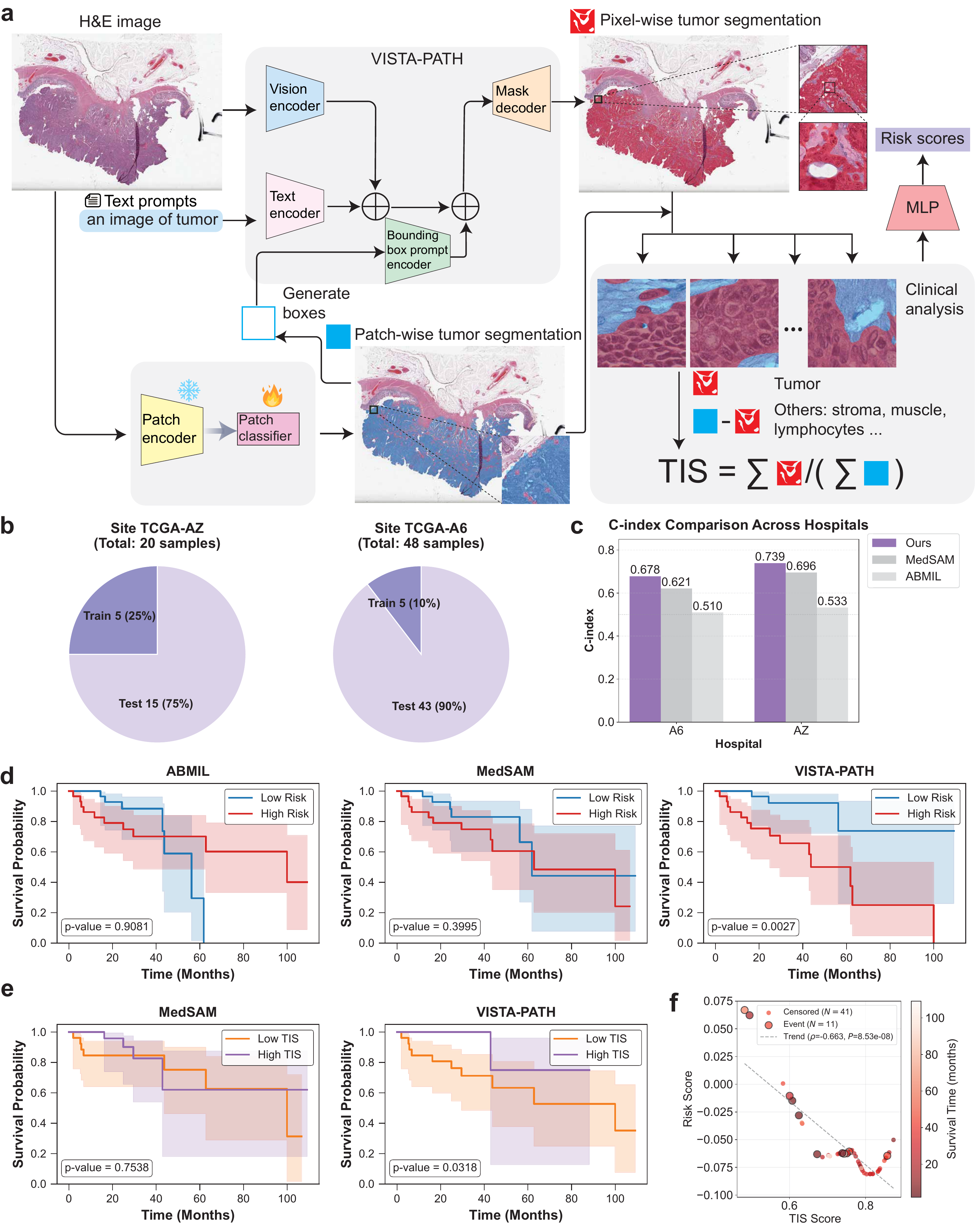}
\end{figure}

\clearpage

\begin{figure}[t] 
\centering
\captionof{figure}{
\textbf{VISTA-PATH enables explainable survival analysis.}
\textbf{a}, Overview of the VISTA-PATH framework for explainable survival analysis on the TCGA-COAD cohort. Given an input whole-slide image, VISTA-PATH produces pixel-wise tumor segmentation, while MUSK generates patch-level predictions. These pixel-wise and patch-wise outputs are combined to compute the Tumor Interaction Score (TIS). TIS is defined as the ratio of the area classified as tumor to the area of all tumor-related classes. The resulting TIS is used as input to a multilayer perceptron (MLP) to predict patient survival outcomes.
\textbf{b}, Data distribution of the training and test splits across the AZ and A6 sites.
\textbf{c}, Comparison of survival prediction performance (C-index) among MedSAM, ABMIL, and VISTA-PATH on the A6 and AZ cohorts.
\textbf{d}, Kaplan--Meier survival curves stratified by the predicted risk scores.
\textbf{e}, Kaplan--Meier survival curves stratified by the proposed Tumor Interaction Score (TIS).
\textbf{f}, Correlation between TIS scores and predicted risk scores. Points are colored by survival time (months), with black-edged markers indicating death events and plain markers indicating censored cases. The dashed line represents the trend, with Spearman's correlation coefficient ($\rho$) and $P$-value shown.
}
\label{fig5:TCGA-COAD}
\end{figure}

\section{Discussion}

In this study, we developed VISTA-PATH as the first interactive and clinically grounded segmentation foundation model for pathology that supports expert reasoning, iterative refinement, and downstream clinical inference. In contrast to prior segmentation foundation models that treat segmentation as a static visual prediction task, VISTA-PATH is designed to resolve extreme multi-tissue heterogeneity, incorporate expert feedback during analysis, and translate segmentation outputs into interpretable clinical signals. VISTA-PATH is trained by the large, ontology-driven VISTA-PATH Data, spanning 93 tissue classes across 9 organs and over 1.6 million masks. We first demonstrate that VISTA-PATH effectively leverages the semantic-spatial modeling for pathology segmentation, highlighting its generalization across diverse tissues and data sources. We then show that VISTA-PATH enables effective human-in-the-loop refinement, allowing limited expert corrections to propagate across whole-slide images with minimal additional annotation effort. Finally, we demonstrate that the segmentation outputs produced by VISTA-PATH can be directly integrated into downstream clinical analyses, enabling the derivation of interpretable and clinically meaningful biomarkers for survival prediction.

VISTA-PATH has several unique strengths.
First, VISTA-PATH effectively leverages the joint semantic--spatial modeling for pathology segmentation. Histopathology images are inherently compositional: tumor epithelium, stroma, immune infiltrates, necrosis, and normal structures are densely interwoven and often visually similar while carrying distinct biological meanings. Accurate pathology segmentation therefore requires models to represent what tissue is present and where it is located at the same time. Existing foundation models decouple these two dimensions: BiomedParse relies on semantic text without explicit spatial grounding, whereas MedSAM relies on bounding boxes without semantic competition between tissue classes. As a result, both struggle when multiple tissue types coexist within overlapping spatial regions. 
VISTA-PATH resolves this limitation by jointly conditioning segmentation on class-aware text prompts and corresponding spatial prompts within a unified vision--language architecture. This design enables simultaneous reasoning over tissue identity and spatial extent, allowing the model to perform true multi-class segmentation in heterogeneous pathology images. The large, ontology-driven VISTA-PATH Data provides the diversity needed to learn this joint representation. The strong and consistent performances across conventional pathology datasets, Visium HD spatial transcriptomics-derived datasets, and Xenium  spatial transcriptomics-derived datasets demonstrate that semantic--spatial coupling generalizes across organs, tissue types, and annotation paradigms. Notably, the largest gains occur in microenvironment-related tissues, where subtle morphology and spatial intermixing dominate, highlighting how joint semantic--spatial modeling directly translates into practical segmentation accuracy.

Second, VISTA-PATH elevates segmentation from a static inference tool to an interactive workflow. The human-in-the-loop experiments show that sparse patch-level corrections, on the order of 10--1,000 patches, can be propagated to whole-slide, pixel-level outputs with high fidelity. This propagation succeeds because VISTA-PATH interprets bounding boxes not merely as geometric constraints, but as class-conditioned spatial signals embedded within its joint semantic--spatial representation.
This capability fundamentally distinguishes VISTA-PATH from prior interactive systems. Traditional approaches require either dense relabeling or full retraining to adapt to new slides, while MedSAM cannot reliably convert patch-level updates into coherent multi-class pixel-wise maps. By contrast, VISTA-PATH transforms limited expert input into global, class-consistent refinements, effectively closing the loop between pathologists and foundation models. This enables continuous refinement during real-world use, rather than one-shot deployment followed by manual correction.

Finally, and most consequentially, VISTA-PATH enables its segmentation outputs to be directly operationalized as clinically meaningful biomarkers for downstream inference. The Tumor Interaction Score (TIS) shows how combining pixel-level tumor maps with patch, level tumor localization yields a quantitative measure of tumor coherence versus infiltration, an established pathology concept that previously required subjective expert judgment. The strong association of TIS with patient survival in two sites from TCGA-COAD cohort, and its ability to outperform existing methods such as ABMIL, demonstrates that segmentation quality and semantic fidelity are not merely aesthetic improvements but have tangible impact on clinical inference. Crucially, when the learning model is removed and TIS alone is used for stratification, VISTA-PATH remains predictive while MedSAM collapses. This shows that the value of TIS arises not from a more powerful classifier, but from biologically faithful segmentation. In summary, VISTA-PATH transforms segmentation foundation models from generic feature extractors into morphological measurement engines capable of producing interpretable biomarkers. This shift, from segmentation as a static prediction to segmentation as a quantitative, clinically grounded representation, fundamentally changes how foundation models can be used in digital pathology.

While VISTA-PATH demonstrates that segmentation can serve as an interactive, research- and clinical-ready foundation for digital pathology, several important challenges remain that point toward future directions of this paradigm. First, although VISTA-PATH enables effective human-in-the-loop refinement through patch-level corrections, pathologists often interact with images using richer cues, including region marking, uncertainty annotation, and iterative hypothesis testing across multiple tissue slides. Extending VISTA-PATH to support these more expressive forms of expert interaction will be essential for deploying foundation models as flexible, collaborative partners in clinical workflows. Second, VISTA-PATH develops Tumor Interaction Score (TIS) illustrating how class-aware segmentation can be converted into an interpretable and clinically meaningful morphological indicator. A key future direction is to move toward automated discovery of morphological biomarkers. With high-fidelity, class-aware segmentation maps as a substrate, foundation models could be used to systematically search the space of spatial tissue configurations, such as tumor--stroma interfaces, immune infiltration patterns, and glandular architecture, to identify new prognostic or predictive signatures that are not predefined by human heuristics. Third, although VISTA-PATH demonstrates strong generalization across organs, datasets, and molecularly derived annotations, further validation across large, multi-institutional clinical cohorts will be required to assess robustness to variations in slide preparation, staining protocols, and scanner hardware. Lastly, while VISTA-PATH can adapt to different slide magnifications, further investigation is needed to determine which magnification levels are most effective for the model, as well as to assess the impact of magnification choices in real-world settings. These factors play a critical role in real research and potential clinical deployment and represent an important next step for translating interactive segmentation foundation models into future use.

\section{Methods} 

\subsection{VISTA-PATH Data: dataset for pathology segmentation}

VISTA-PATH Data was curated through the systematic integration of 22 publicly available segmentation resources from multiple platforms, including The Cancer Imaging Archive (TCIA), Kaggle, Grand Challenge, Scientific Data, CodaLab, and segmentation challenges hosted by the Medical Image Computing and Computer Assisted Intervention Society (MICCAI). The sources of all datasets are listed in \textbf{Extended Table~\ref{sup:internal_data_source}}. Each image is associated with one or more tissue types and each tissue type is annotated with a high-precision segmentation mask and a canonical semantic label derived from a tissue ontology of structured pathology. This comprehensive dataset establishes a unified foundation for advancing segmentation models in computational pathology by capturing a broad spectrum of tissue and object diversity across multiple organs and imaging contexts.

To ensure the quality and consistency of VISTA-PATH Data, we applied stringent inclusion criteria: each image was required to be manually or semi-manually segmented at the pixel level, and each segmented object had to be associated with a clearly defined name in the original dataset description. Given the diversity of data sources and formats, we standardized all inputs into a unified representation through the following processing steps:

\begin{itemize}
\item \textbf{Image format transformation.} Pathology images were originally provided in diverse resolutions and formats, including both patch-wise and whole-slide image (WSI) levels, and were stored in heterogeneous file types such as \texttt{.tif}, \texttt{.png}, and \texttt{.svs}. To ensure interoperability and facilitate downstream processing, all images were converted into a standardized three-channel format with a consistent structure and naming convention. For WSI data, the highest resolution level was selected to preserve the original image quality and spatial details.

\item \textbf{Mask format transformation.} The pixel-wise annotation masks were originally provided in a variety of heterogeneous formats. Some datasets stored the masks directly as raster images (\textit{e.g.}, \texttt{.png}, \texttt{.tif}), whereas others represented annotations as polygon coordinates in \texttt{.csv} or \texttt{.json} files. To unify these representations, all polygon-based annotations were rasterized into pixel-level segmentation masks that are spatially aligned with their corresponding pathology images. Formally, let $\Omega \subset \mathbb{R}^2$ denote the image domain and $\{P_c\}_{c=1}^C$ represent the polygonal regions for each class $c$. The rasterized mask $M : \Omega \rightarrow \{0, 1, \dots, C\}$ is defined as

\begin{equation}
M(p) =
\begin{cases}
c, & p \in P_c \\
0, & p \notin \bigcup_{c=1}^C P_c,
\end{cases}
\end{equation}

where $p = (x, y)$ denotes a pixel coordinate, and $c$ is the class index (0 corresponds to background). Each mask was then encoded as either binary or multi-class images, depending on the number of annotated structures, ensuring that each pixel is assigned to a unique semantic class. This standardized mask format enables consistent downstream processing and model training across all datasets in VISTA-PATH Data.

\item \textbf{Label class construction.} We manually extracted and verified the correctness of each class label to ensure consistency across datasets. The detailed class definitions are summarized in \textbf{Fig.~\ref{fig:overview}c} and \textbf{Extended Data Fig.~\ref{sup:fig1}}. For datasets DROV and DROID, class names were extracted from the descriptions and metadata of the data set. We then performed manual filtering and standardization to remove redundant or ambiguous labels. Each validated class was subsequently mapped to its corresponding region in the segmentation mask, ensuring a one-to-one correspondence between semantic labels and pixel-level annotations.

\end{itemize}

For each dataset, we randomly split the samples into 95\% for training and 5\% for testing, and the training--testing splits are conducted at the sample-level. To mimic real-world usage scenarios in which pathology images are typically acquired at different magnifications, we did not restrict the dataset to a single magnification level. To address the variability in image sizes across datasets, all whole-slide or patch-level images were cropped into patches of size $1024 \times 1024$ pixels. During model training, a random $512 \times 512$ region was further sampled from each patch and subsequently resized to $224 \times 224$ pixels before being fed into the network. The same sampling and resizing strategy was applied to the corresponding segmentation masks to ensure spatial alignment between the images and their annotations.

\subsection{External Datasets}

To further evaluate the generalization capability of our model, we conducted additional experiments on several external datasets that were not seen during the training phase. These datasets were collected from independent sources and represent different imaging conditions, organs, and tissue types compared to the training data. By evaluating the model on unseen datasets, we aim to assess its robustness and transferability to real-world scenarios, where domain shifts in acquisition protocols and biological variability commonly occur. The sources of all external datasets are listed in \textbf{Extended Table~\ref{sup:external_data_source}}.

\subsubsection{Pathology Segmentation Datasets}

Two additional publicly available pathology segmentation datasets LungHP and OCDC were used to evaluate the generalizability of our models. The images, masks, and class labels from these datasets were processed using the same preprocessing pipeline described above for VISTA-PATH Data.

\subsubsection {Visium HD Datasets}
We collected five Visium HD datasets from the 10x Genomics platform, including Colon, Kidne, Lung, Pancreas, and Prostate. Image statistics are presented in \textbf{Fig~\ref{fig3:external_results}b}. Each Visium HD slide is represented as a set of $N$ spatial bins $\{b_i\}_{i=1}^N$, where each bin $b_i$ corresponds to a local region containing a molecular expression profile $\mathbf{x}_i \in \mathbb{R}^d$.

To obtain semantically meaningful tissue regions, we used $k$-means clustering pipeline~\cite{ahmed2020k} provided by 10X Genomics platform to group bins with similar molecular features. Formally, $k$-means minimize the within-cluster variance:
\begin{equation}
\underset{\{C_j\}_{j=1}^K}{\arg\min} 
\sum_{j=1}^{K} \sum_{\mathbf{x}_i \in \mathcal{C}_j} 
\|\mathbf{x}_i - \boldsymbol{\mu}_j\|_2^2,
\end{equation}
where $C_j$ denotes the $j$-th cluster and $\boldsymbol{\mu}_j$ is its centroid. Each bin $b_i$ is assigned to the cluster
\begin{equation}
c_i = \underset{j \in \{1,\dots,K\}}{\arg\min} \|\mathbf{x}_i - \boldsymbol{\mu}_j\|_2^2.
\end{equation}

To connect clusters with segmentation textual class description, pathologists from Penn Medicine then manually examined the clusters $\{C_j\}$ on each slide to verify whether the grouping corresponded to meaningful tissue structures or biological patterns. After annotation, each cluster is assigned a semantic class label. To generate a pixel-level segmentation mask, we leveraged the known bin size provided by the Visium HD platform (\textit{i.e.}, \(8~\mu\mathrm{m} \times 8~\mu\mathrm{m}\) per bin). For each bin, all pixels located within its spatial boundary were assigned the same class label as the cluster label. This bin-to-pixel mapping procedure converts cluster-level annotations into dense pixel-level labels, resulting in a segmentation dataset.

\subsubsection{Xenium Datasets}
We further collected 22 Xenium datasets from the 10x Genomics platform. Compared to Visium HD with bin-level aggregated spatial transcriptomic information, Xenium offers a precise single-cell resolution with explicit cell segmentation contours and corresponding gene profiles. For each Xenium slide, we first obtained the cell segmentation contours provided by the platform, where each cell corresponds to a segmented region and is associated with a gene expression profile $\mathbf{z}_i \in \mathbb{R}^d$. 
To cluster cells with similar gene expression profiles, we applied CellCharter~\cite{varrone2024cellcharter} to compute multi-layer spatially aggregated cell embeddings, in which each cell is represented by a vector that jointly encodes its intrinsic transcriptional profile and the molecular composition of its local neighborhood. These embeddings were then clustered using $K$-means to group cells with similar molecular states into discrete spatial domains. The resulting domain labels were subsequently projected onto the co-registered H\&E images for tissue segmentation label assignment.

\paragraph{Spatial domain modeling via CellCharter-based neighborhood embeddings and K-means clustering.}
Let $x_i \in \mathbb{R}^d$ denote the PCA-reduced gene expression profile of cell $i$, and let $\mathcal{G}=(V,E)$ be the spatial adjacency graph constructed over all cells using Delaunay triangulation on their spatial coordinates. We employed CellCharter's neighborhood aggregation operator to compute spatially contextualized embeddings by recursively aggregating features from multi-hop neighborhoods on $\mathcal{G}$. Specifically, starting from the initial representation $X^{(0)} = [x_1, \ldots, x_N]$, we generated a sequence of aggregated features
\begin{equation}
X^{(l+1)} = \mathrm{Agg}(X^{(l)}, \mathcal{G}), \quad l = 0, \ldots, L-1,
\end{equation}
where $\mathrm{Agg}(\cdot)$ denotes CellCharter's graph-based neighbor pooling. The final embedding $Z = [X^{(0)}, X^{(1)}, \ldots, X^{(L)}]$ concatenates information from multiple spatial scales, capturing both single-cell transcriptional states and progressively larger molecular neighborhoods. In our experiments, we used $L=3$ aggregation layers, resulting in a multi-scale spatial embedding for each cell.

We then performed unsupervised clustering on these CellCharter-derived embeddings $Z_i$ using MiniBatch K-means. Given a specified number of clusters $K$, K-means optimizes
\begin{equation}
\min_{\{c_i\}, \{\mu_k\}} \sum_{i=1}^{N} \left\lVert Z_i - \mu_{c_i} \right\rVert^2,
\end{equation}
where $c_i \in \{1, \ldots, K\}$ is the cluster assignment for cell $i$ and $\mu_k$ denotes the centroid of cluster $k$. This produces hard assignments of cells to spatial domains, yielding a discrete partitioning of the tissue into molecularly defined regions. Finally, the cluster labels were projected onto the co-registered H\&E whole-slide image using the Xenium-to-histology coordinate transformation.

After the clusters are obtained, we invite pathologists from Penn Medicine annotate each cluster with tissue classes with a textual description. The pixel-level segmentation labels are obtained by assigning each pixel within a cell's contour the class name of the cluster to which that cell belongs. In summary, the cell-to-pixel assignment from Xenium yields a dense and precise semantic segmentation mapping.

\subsection{Details of VISTA-PATH model architecture}

VISTA-PATH is a visual interactive segmentation foundation model for pathology tissue analysis that supports both interactive visual prompts and text prompts. Specifically, VISTA-PATH is composed of four key components: an image encoder, a text encoder, a bounding box prompt encoder, and a mask decoder (\textbf{Fig.~\ref{fig:overview}}). The functionality and design of each module are described in detail below.

\paragraph{Image encoder.}
We adopt the vision encoder from the pretrained PLIP model to extract image embedding for input pathology images. 
Given an input image $I \in \mathbb{R}^{3 \times 224 \times 224}$, the encoder produces a set of patch embeddings $\mathbf{V}'=\{v_i'\}_{i=1}^N$ with $\mathbf{V}'\in\mathbb{R}^{N\times d'}$, where $N=49$ corresponds to a $7\times7$ patch grid and $d'=768$ is the original embedding dimension. 
After removing the class token, the remaining patch embeddings are projected through a visual projection layer into a shared latent space:
\begin{equation}
\mathbf{V}=\mathbf{V}'\,\mathbf{W}_{\mathrm{proj}}, 
\qquad 
\mathbf{W}_{\mathrm{proj}}\in\mathbb{R}^{d'\times d},\;
\mathbf{V}\in\mathbb{R}^{N\times d},\;
d=512.
\end{equation}
This yields semantical embeddings which will be utilized for segmentation predictions.

\paragraph{Text encoder.}
We adopt the PLIP text encoder to process textual class description (\textit{e.g.}, ``an image of \texttt{\{class\_name\}}''). 
The input text prompt is tokenized into a sequence of $T$ tokens and encoded as $\mathbf{T}'=\{t_j'\}_{j=1}^{T}$ with $\mathbf{T}'\in \mathbb{R}^{T\times d'}$, where $T=77$ and $d'=768$ is the original textual embedding dimension. 
These embeddings are then projected into the shared latent space used for cross-modal fusion:
\begin{equation}
\mathbf{T}=\mathbf{T}'\,\mathbf{W}_{\mathrm{text}}, 
\qquad 
\mathbf{W}_{\mathrm{text}}\in\mathbb{R}^{d'\times d},\;
\mathbf{T}\in\mathbb{R}^{T\times d},\;
d=512.
\end{equation}
This yields text representations aligned with the image embedding space, enabling subsequent cross-modal interactions.

\paragraph{Prompt encoder.} 
We integrate a frozen prompt encoder from the SAM. A bounding box prompt $B \in \mathbb{R}^{1 \times 4}$ is encoded into a prompt embedding
\begin{equation}
\mathbf{P} = \{ p \}, \quad \mathbf{P} \in \mathbb{R}^{2 \times d}, d=512.
\end{equation}

\paragraph{Cross-modal fusion and context modeling.} 
We employ cross-attention modules to fuse visual, textual, and spatial prompt information. 
Given image embeddings $\mathbf{V}\!\in\!\mathbb{R}^{N\times d}$, text embeddings $\mathbf{T}\!\in\!\mathbb{R}^{T\times d}$, and a bounding-box prompt embedding $\mathbf{P}\!\in\!\mathbb{R}^{2\times d}$ (with $N=7 \times 7$, $T=77$, $d=512$ as defined above), fusion proceeds in two stages:

\begin{equation}
\mathbf{F}_{\text{text}}=\mathrm{CrossAttn}\big(\mathbf{Q}=\mathbf{V},\;\mathbf{K}=\mathbf{T},\;\mathbf{V}=\mathbf{T}\big),
\qquad
\mathbf{F}_{\text{prompt}}=\mathrm{CrossAttn}\big(\mathbf{Q}=\mathbf{F}_{\text{text}},\;\mathbf{K}=\mathbf{P},\;\mathbf{V}=\mathbf{P}\big).
\end{equation}

The first stage lets each visual token attend to semantically relevant textual cues, while the second stage injects localized spatial priors from the prompt embedding, yielding a unified representation for downstream decoding.

The fused representation $\mathbf{F}_{\text{prompt}}$ is further refined by a stack of Transformer encoder layers~\cite{vaswani2017attention}:
\begin{equation}
\mathbf{F}_{\text{final}} = \mathrm{TransformerEncoderLayer}^L(\mathbf{F}_{\text{prompt}}),
\end{equation} 
where $L = 4$, $\mathbf{F}_{\text{final}}\in\!\mathbb{R}^{512\times 7\times 7}$.

\paragraph{Mask decoder.}
Given the fused feature map $\mathbf{F}_{\text{final}}\in\!\mathbb{R}^{512\times 7\times 7}$, 
the mask decoder produces dense per-pixel predictions 
$\hat{\mathbf{Y}}\in\mathbb{R}^{K\times 224\times 224}$ via a 
four-stage upsampling stack. Each stage applies bilinear upsampling 
followed by a $3{\times}3$ convolution, Batch Normalization, and ReLU, 
progressively refining spatial details while expanding the receptive field. 
Concretely, the resolution evolves as $7{\to}14{\to}28{\to}56{\to}224$, with 
channel widths $\{512{\to}256{\to}128{\to}64{\to}32\}$. A final $1{\times}1$ 
convolution maps features to $K$ class logits per pixel. 

Formally, letting $U(\cdot)$ denote bilinear upsampling and 
$\phi(\cdot)$ denote $\mathrm{Conv}(3{\times}3)$--BN--ReLU, the decoder is
\begin{equation}
\mathbf{H}_1 = U\!\big(\phi(\mathbf{F}_{\text{final}})\big),\quad
\mathbf{H}_2 = U\!\big(\phi(\mathbf{H}_1)\big),\quad
\mathbf{H}_3 = U\!\big(\phi(\mathbf{H}_2)\big),\quad
\mathbf{H}_4 = U\!\big(\phi(\mathbf{H}_3)\big),
\end{equation}
\begin{equation}
\hat{\mathbf{Y}}=\mathrm{Conv}_{1\times 1}(\mathbf{H}_4)\in\mathbb{R}^{K\times 224\times 224}.
\end{equation}
In our experiments we set $K=2$ (foreground and background), producing a binary segmentation for each image--text pair corresponding to the specified text prompt.

\subsection{Details of VISTA-PATH model training} 

VISTA-PATH is trained for grounded segmentation using text prompts, optionally grounded with bounding-box prompts as spatial priors. 
For each image--text pair, the ground-truth mask is defined by the prompt: pixels belonging to the specified class are labeled 1 and others 0, \textit{i.e.}, $\mathbf{Y}\in\{0,1\}^{H\times W}$. 
The model outputs two logits per pixel $\hat{\mathbf{Y}}\in\mathbb{R}^{2\times H\times W}$ (background/foreground). 
Let $\mathbf{p}_{\cdot,ij}=\mathrm{softmax}\big(\hat{\mathbf{Y}}_{\cdot,ij}\big)\in[0,1]^2$ denote the per-pixel class probabilities and write $p_{\mathrm{fg},ij}$ and $p_{\mathrm{bg},ij}$ for the foreground and background entries, respectively. 
We minimize the softmax cross-entropy~\cite{mao2023cross}:
\begin{equation}
\mathcal{L}_{\mathrm{CE}}
= -\,\frac{1}{HW}\sum_{i=1}^{H}\sum_{j=1}^{W}
\Big[\,Y_{ij}\,\log p_{\mathrm{fg},ij} + (1-Y_{ij})\,\log p_{\mathrm{bg},ij}\Big].
\end{equation}
When a bounding-box prompt is available, it is used as an additional spatial prior during forward propagation; the objective remains the same.

We initialize the VISTA-PATH image and text encoders from the PLIP model. Training uses mixed-precision (FP16) with the text encoder frozen. We optimize the model parameters with a learning rate of $5\times10^{-5}$ for 10 epochs using a batch size of 512. Training is performed on a single NVIDIA H200 GPU and completes in approximately 25 hours. For evaluation, inference is run on one NVIDIA H200 GPU with a memory usage of about 2.4\,GB.

\subsection{Evaluation metrics and statistical analysis}

We assess segmentation using the Dice coefficient~\cite{dice1945measures}. For image $n$ with prediction $\hat{Y}_n$ and ground truth $Y_n$,
\begin{equation}
\mathrm{Dice}_n \;=\; \frac{2\,|\hat{Y}_n \cap Y_n|}{|\hat{Y}_n| + |Y_n|}.
\end{equation}
For binary tasks, we compute $\mathrm{Dice}_n$ per image and report mean over the test set. For multi-class tasks, we compute Dice per class and report the average across classes. Following common practice, if both $\hat{Y}_n$ and $Y_n$ are empty we set $\mathrm{Dice}_n{=}1$; if only one is empty we set $\mathrm{Dice}_n{=}0$.

\paragraph{Statistical testing.}

We report per-image Dice and summarize each method's performance by the mean and its 95\% confidence interval computed via percentile bootstrap (10{,}000 resamples).
Statistical significance was assessed using a two-sided Student's $t$-test on per-image Dice score differences. Statistical significance is annotated as $^{*}P < 0.05$; $^{**}P < 1 \times 10^{-2}$; $^{***}P < 1 \times 10^{-3}$. $P$ value in the Kaplan–Meier~\cite{se1958non} survival plot was computed using the log-rank test with a chi-square approximation~\cite{lee2003statistical}.

\subsection{Details of experiments on internal segmentation evaluation}

We report per-dataset results for every dataset in the internal benchmark, with the macro mean (averaged across datasets) summarized in \textbf{Extended Data Table~\ref{sup:table_internal}}.
In the pathology domain, it is crucial to have a model capable of handling different organs. To assess performance across anatomical contexts, we aggregate results by organ. For each organ, we compute the mean Dice score with 95\% percentile-bootstrap confidence intervals (10{,}000 resamples) over all images belonging to datasets of that organ.
Another important consideration is the diversity of pathology classes and how well a segmentation model handles them. To assess class-wise performance, we standardize labels into three categories: tumor-related, microenvironment-related, and normal anatomical tissues (see \textbf{Fig.~\ref{fig2:internal_results}}). Per-category scores are obtained by pooling images whose ground-truth foreground belongs to the corresponding group and then computing the mean Dice score.

\subsection{Details of experiments on external segmentation evaluation}

To further assess the generalizability of VISTA-PATH, we evaluate the model on external datasets not seen during training. In the zero-shot setting, no fine-tuning or dataset-specific adaptation is performed; we use the same inference protocol and prompts across datasets. We report per-image Dice scores. A summary of the external datasets (domains, organs, modalities, and class definitions) is provided in \textbf{Extended Data Fig.~\ref{sup:fig3}}.

Similar to the evaluation of internal datasets, we also report organ-wise and class-wise results. For class-wise evaluation, labels are standardized into three categories: tumor-related, microenvironment-related, and normal anatomical tissues, and we report the mean Dice. Organ-wise performance is obtained by aggregating images by organ and computing the corresponding mean Dice.

\subsection{Details of experiments on Human-in-the-loop study}

In practice, segmentation should not be treated as a one-time inference. Pathologists routinely review model outputs and iteratively refine them to correct errors and improve accuracy. Crucially, such corrections should be supported at the whole-slide image level, as whole-slide images are essential for reliable downstream analysis. Motivated by this need, we propose a human-in-the-loop framework that leverages VISTA-PATH as an interactive segmentation model to iteratively refine whole-slide image segmentation results.

An overview of the human-in-the-loop framework is shown in \textbf{Fig.~\ref{fig4:human-in-the-loop}a}. Given a whole-slide image as input, the image is first tiled into fixed-size patches. A MUSK model~\cite{xu2021predicting} is then applied to extract patch embeddings and generate patch-level classification predictions, which serve as the initial segmentation output. Pathologists review these patch-level results and provide localized corrections on selected subregions. The corrected patch-level annotations are collected and used to train a lightweight classifier (\textit{i.e.}, an XGBoost head~\cite{chen2016xgboost}) on the corresponding extracted patch embeddings. The updated patch classifier is subsequently applied to all patches to produce refined patch-level segmentation results. The refined patch-level predictions are then used to derive bounding-box prompts for VISTA-PATH. Specifically, we apply a sliding-window strategy over the whole-slide image; for each window, a tight bounding box is computed from the local binary mask and provided as a prompt to VISTA-PATH, enabling refined pixel-wise segmentation.

The iterative refinement process is terminated once pathologists judge the segmentation results to be satisfactory. The number of iterative rounds and the number of patch-level annotations provided at each round are summarized in \textbf{Fig.~\ref{fig4:human-in-the-loop}b--c} and \textbf{Extended Data Tables~\ref{tab:dice_visium}--\ref{tab:dice_xenium}}.

\subsection{Details of experiments on TCGA-COAD cohort study}

Explainable AI is essential for enhancing the interpretability and trustworthiness of pathology-based clinical models. 
In pathology, accurate prediction of cancer survival is fundamentally based on the identification, morphological assessment, and quantitative analysis of specific pathological structures, such as tumor regions. Consequently, precise segmentation of key regions and reliable extraction of morphological measures are critical for generating meaningful survival signals. In this study, we demonstrate how VISTA-PATH improves survival analysis in the TCGA-COAD cohort by providing accurate segmentation masks and enabling the extraction of robust morphological measurements, which directly improves survival predictions.

We introduce the \textbf{Tumor Interaction Score (TIS)}, a metric designed to quantify tumor solidity and assess the degree of tumor interacting with other microenvironments. A more solid tumor component often reflects a less invasive phenotype, slower disease progression, or earlier disease stage, all of which are associated with more favorable clinical outcomes. TIS is computed through a two-stage process: we first generate patch-level tumor segmentations and then perform pixel-level segmentation within the predicted tumor patches. The score is defined as the ratio between the intersection of pixel- and patch-level tumor masks and the total patch-level tumor area:
\begin{equation}
\mathrm{TIS} = \frac{\sum_{i=1}^{N} \left| P_i \cap S \right|}{\sum_{i=1}^{N} \left| P_i \right|},
\end{equation}
where $P_i$ denotes the $i$-th patch classified as tumor by the patch-level predictor, $S$ denotes the pixel-wise tumor segmentation mask, and $\lvert \cdot \rvert$ indicates area measured in pixels. This metric quantifies the density and cohesiveness of the tumor component, with higher values indicating more well-demarcated solid tumor regions that correspond to reduced invasiveness and improved patient outcomes. 

We compare VISTA-PATH with two categories of baselines: a standard multiple instance learning (MIL) approach (\textit{i.e.}, ABMIL) and segmentation-based methods (\textit{i.e.}, MedSAM). For the MIL baseline, ABMIL operates on patch embeddings extracted using the MUSK model. For the segmentation-based baseline, we replace VISTA-PATH with MedSAM and compute the corresponding TIS score for survival prediction. Under segmentation-based methods, each whole-slide image (WSI) yields a single TIS score, which is fed into a multilayer perceptron (MLP) to generate the final risk prediction.

\section*{Acknowledgments}

We would like to thank Zhengde (Theodore) Zhao for providing valuable insights on text prompt-based image segmentation.

\section*{Data Availability}
\textbf{DROID} data used in this study were obtained from the Analytic Imaging Diagnostics Arena (AIDA). 
The dataset is released under a permissive research license that allows use, modification, and redistribution for medical diagnostics research, provided that the original copyright and permission notices are retained. 
Data are available at \url{https://datahub.aida.scilifelab.se/10.23698/aida/drsk}. \textbf{DROV} dataset is free for use in legal and ethical medical diagnostics research. 
Access to the data requires approval through the AIDA data portal. 
The designated recipient researcher must hold a PhD degree in a relevant field, and the applicant must be an authorized institutional signatory with the legal authority to enter into data sharing agreements on behalf of the institution. Data are available at \url{https://datahub.aida.scilifelab.se/10.23698/aida/drov}. 
\textbf{KPMP} data is supported by the National Institute of Diabetes and Digestive and Kidney Diseases (NIDDK) through grants 
U01DK133081, U01DK133091, U01DK133092, U01DK133093, U01DK133095, U01DK133097, 
U01DK114866, U01DK114908, U01DK133090, U01DK133113, U01DK133766, U01DK13\\3768, 
U01DK114907, U01DK114920, U01DK114923, U01DK114933, U24DK114886, UH3DK114926, 
UH3DK114861, UH3DK114915, and UH3DK114937. We gratefully acknowledge the essential 
contributions of patient participants and the support of the American public 
through their tax dollars. Data are available at \url{https://www.kpmp.org/available-data}. \textbf{TCGA-COAD} results here are in whole or part based upon data generated by the TCGA Research Network: \url{https://www.cancer.gov/tcga}. All data resources used in this study are summarized in \textbf{Extended Data Table~\ref{sup:internal_data_source}} and \textbf{Extended Data Table~\ref{sup:external_data_source}}.

\section*{Code Availability}
VISTA-PATH (new software developed in this study) is released as an open-source Python library on GitHub at \url{https://github.com/zhihuanglab/VISTA-PATH}.

\section*{Ethics declarations}
\subsection*{Competing interests}
Yucheng Tang and Daguang Xu are employees at the Nvidia Corporation, USA.

\section*{Author contributions statement}

\textbf{Conceptualization}: Peixian Liang, Zhi Huang. \textbf{Methodology}: Peixian Liang, Songhao Li, Shunsuke koga, Yutong Li, Zhi Huang. \textbf{Experiments}: Peixian Liang, Songhao Li, Shunsuke koga, Yucheng Tang, Daguang Xu, Zhi Huang. \textbf{Data curation and acquisition}: Peixian Liang, Yutong Li, Zhi Huang. \textbf{Expert Pathology Annotation, Interpretation, and Feedback}: Shunsuke Koga, Zahra Alipour. \textbf{Manuscript writing}: Peixian Liang, Zhi Huang. \textbf{Supervision}: Zhi Huang. All authors reviewed and approved the final manuscript.

\bibliography{sample}

\clearpage
\setcounter{figure}{0}
\renewcommand{\thefigure}{\arabic{figure}}

\setcounter{table}{0}
\renewcommand{\thetable}{\arabic{table}}

\begin{figure}[tbph]
\captionsetup{labelformat=empty}
\centering
\includegraphics[width=\textwidth]{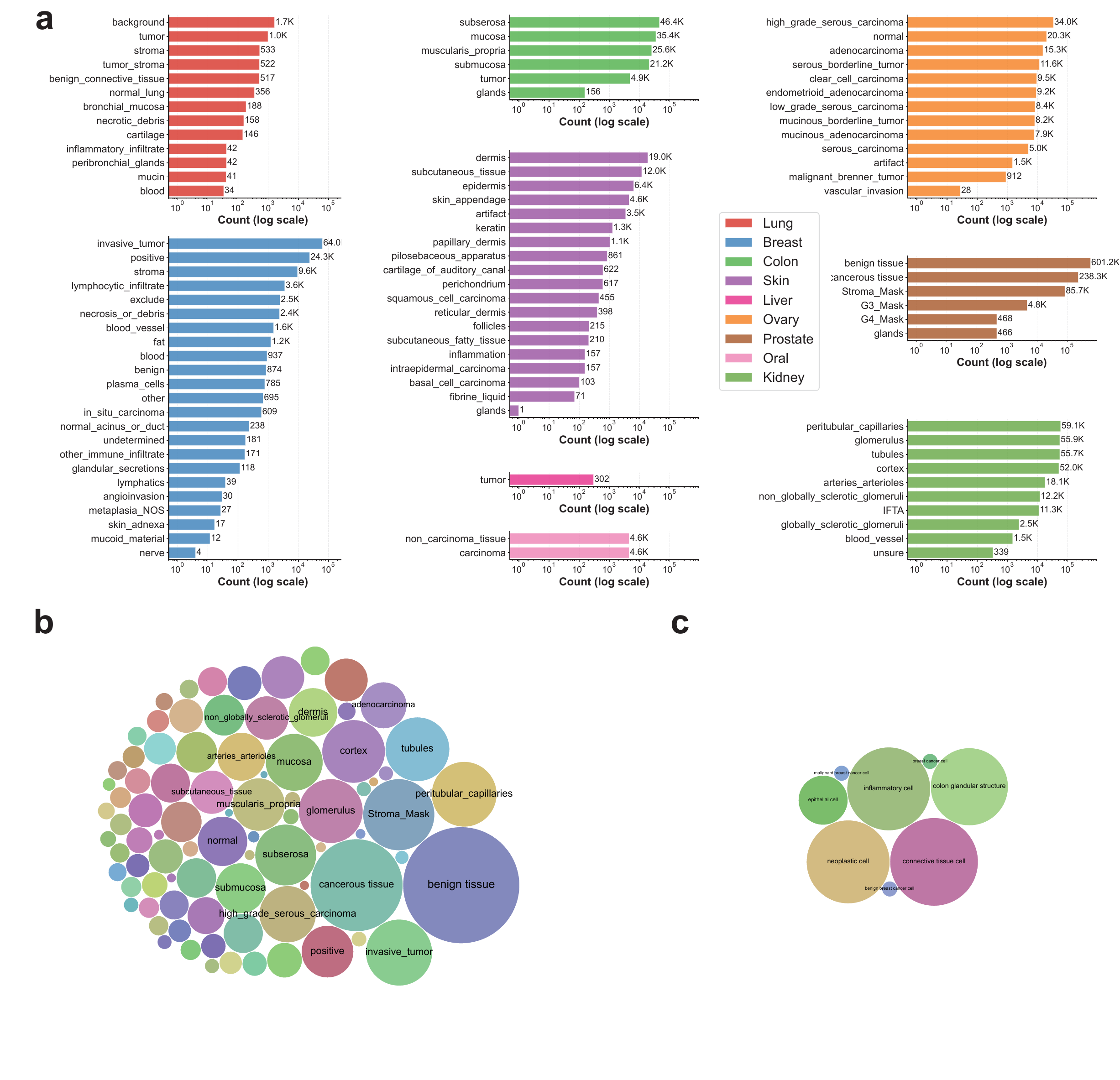}
\caption{\textbf{Extended Data Figure~1. Overview of VISTA-PATH Data.} 
\textbf{a}, Bar plots showing the distribution of tissue classes across organs in the VISTA-PATH Dataset, grouped by organ type. Counts are displayed on a logarithmic scale.
\textbf{b}, Bubble plot summarizing the pathology tissue classes in VISTA-PATH Data, where each bubble represents a tissue class and its size is proportional to the number of annotated image--mask instances.
\textbf{c}, Bubble plot of pathology tissue class distribution in the BiomedParse dataset, shown using the same visualization scheme as in \textbf{b}, Compared with VISTA-PATH, BiomedParse exhibits substantially fewer tissue categories and a more limited semantic coverage. }
\label{sup:fig1}
\end{figure}

\begin{figure}[tbph]
\captionsetup{labelformat=empty}
\centering
\includegraphics[width=\textwidth]{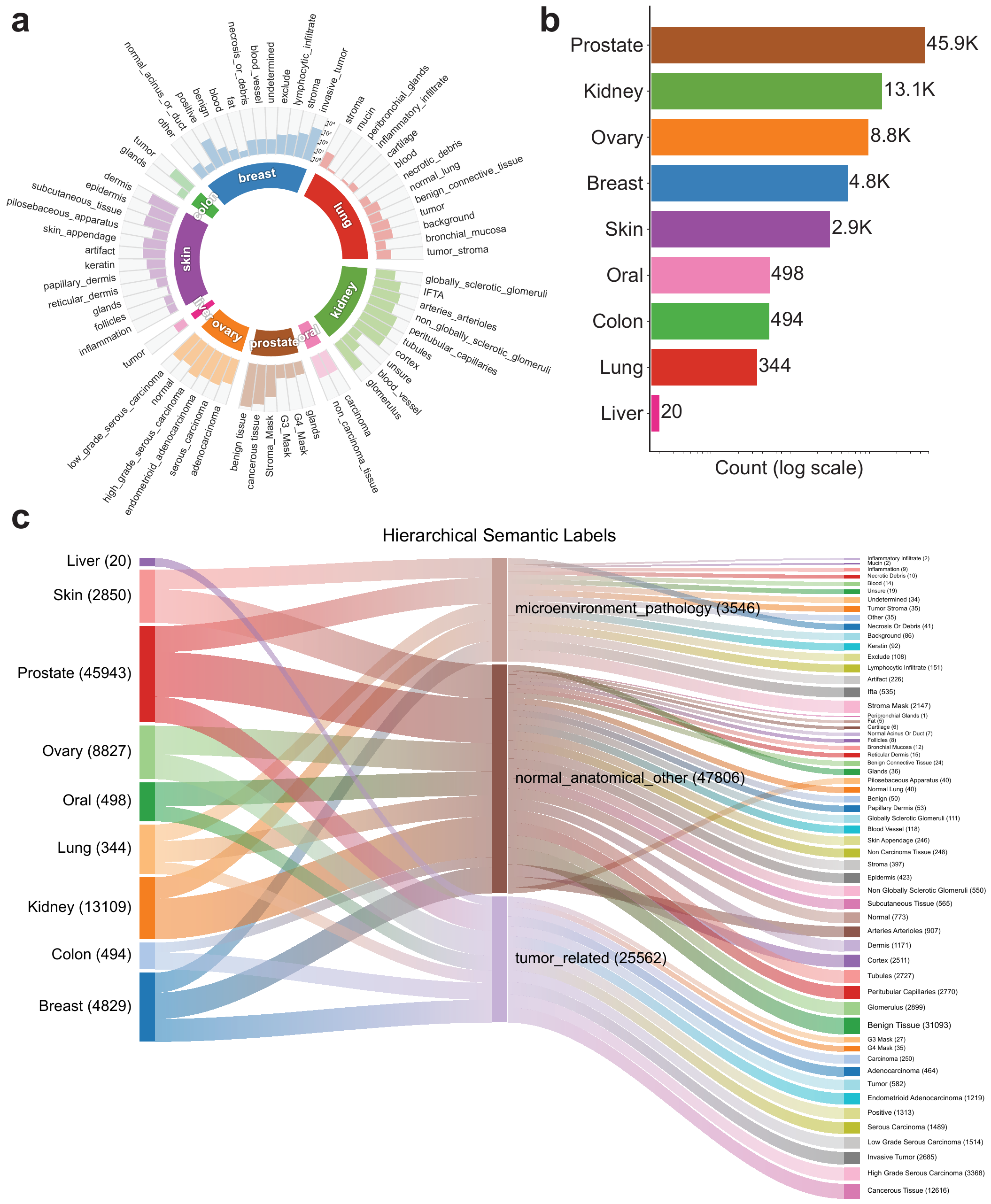}
\caption{\textbf{Extended Data Figure~2. Overview of the held-out internal datasets used for evaluation.}
\textbf{a}, Sunburst diagram illustrating the hierarchical organization of tissue classes in the held-out internal evaluation datasets. Inner rings denote organ types, while outer rings correspond to pathology tissue labels.
\textbf{b}, Bar plot showing the number of annotated image-mask instances for each organ in the held-out internal evaluation datasets.
\textbf{c}, Sankey diagram visualizing the hierarchical semantic structure of tissue types in the held-out internal datasets. Flow width is proportional to the number of samples, with sample counts shown in parentheses. }
\label{sup:fig2}
\end{figure}

\begin{figure}[tbph]
\captionsetup{labelformat=empty}
\centering
\includegraphics[width=\textwidth]{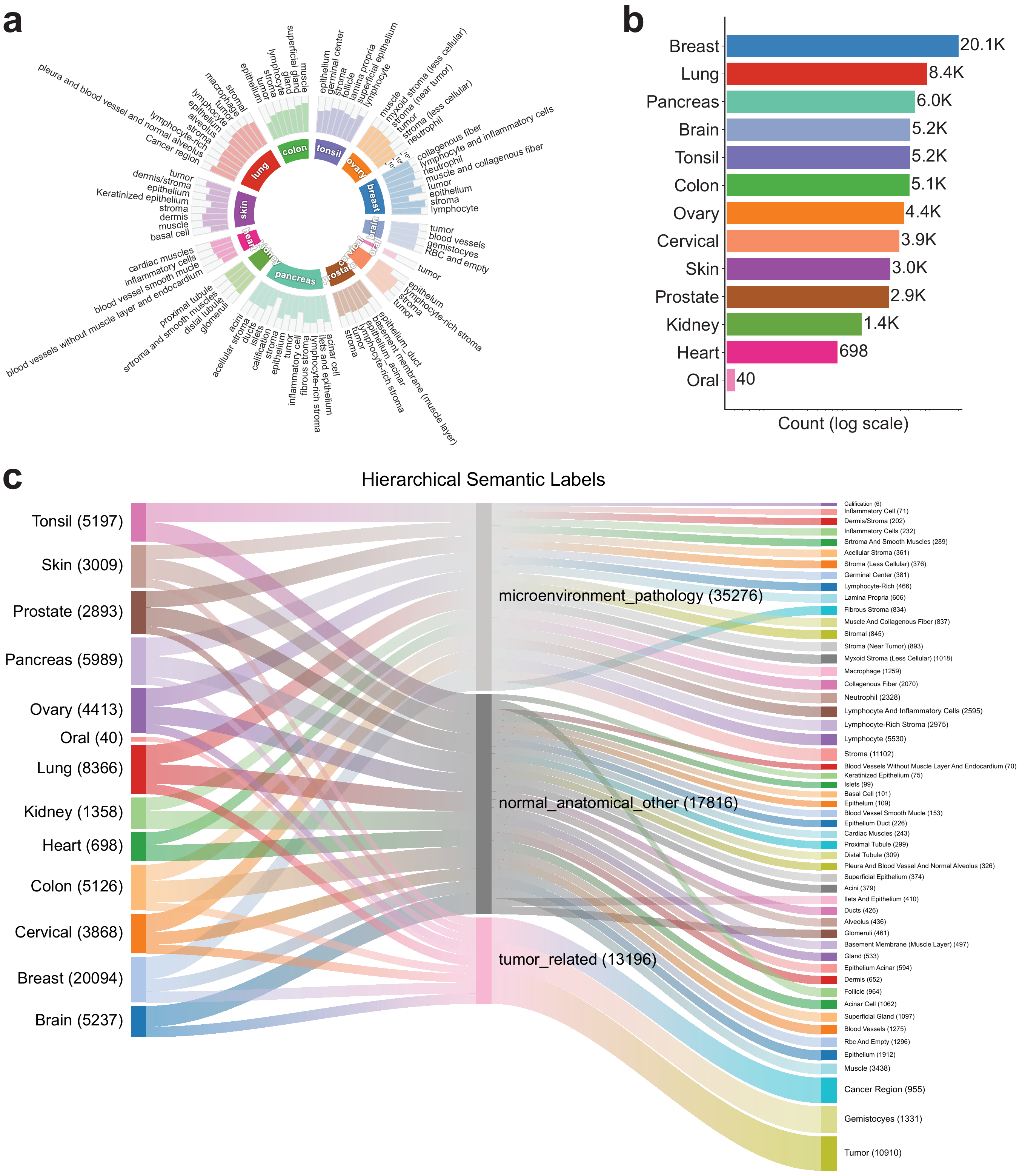}
\caption{\textbf{Extended Data Figure~3. Overview of the external datasets used for evaluation.}
\textbf{a}, Sunburst diagram illustrating the hierarchical organization of tissue classes in the external evaluation datasets. Inner rings denote organ types, while outer rings correspond to pathology tissue labels.
\textbf{b}, Bar plot showing the number of annotated image-mask instances for each organ in the external evaluation datasets.
\textbf{c}, Sankey diagram visualizing the hierarchical semantic structure of tissue types in the external datasets. Flow width is proportional to the number of samples, with sample counts shown in parentheses.}
\label{sup:fig3}
\end{figure}

\begin{figure}[tbph]
\captionsetup{labelformat=empty}
\centering
\includegraphics[width=\textwidth]{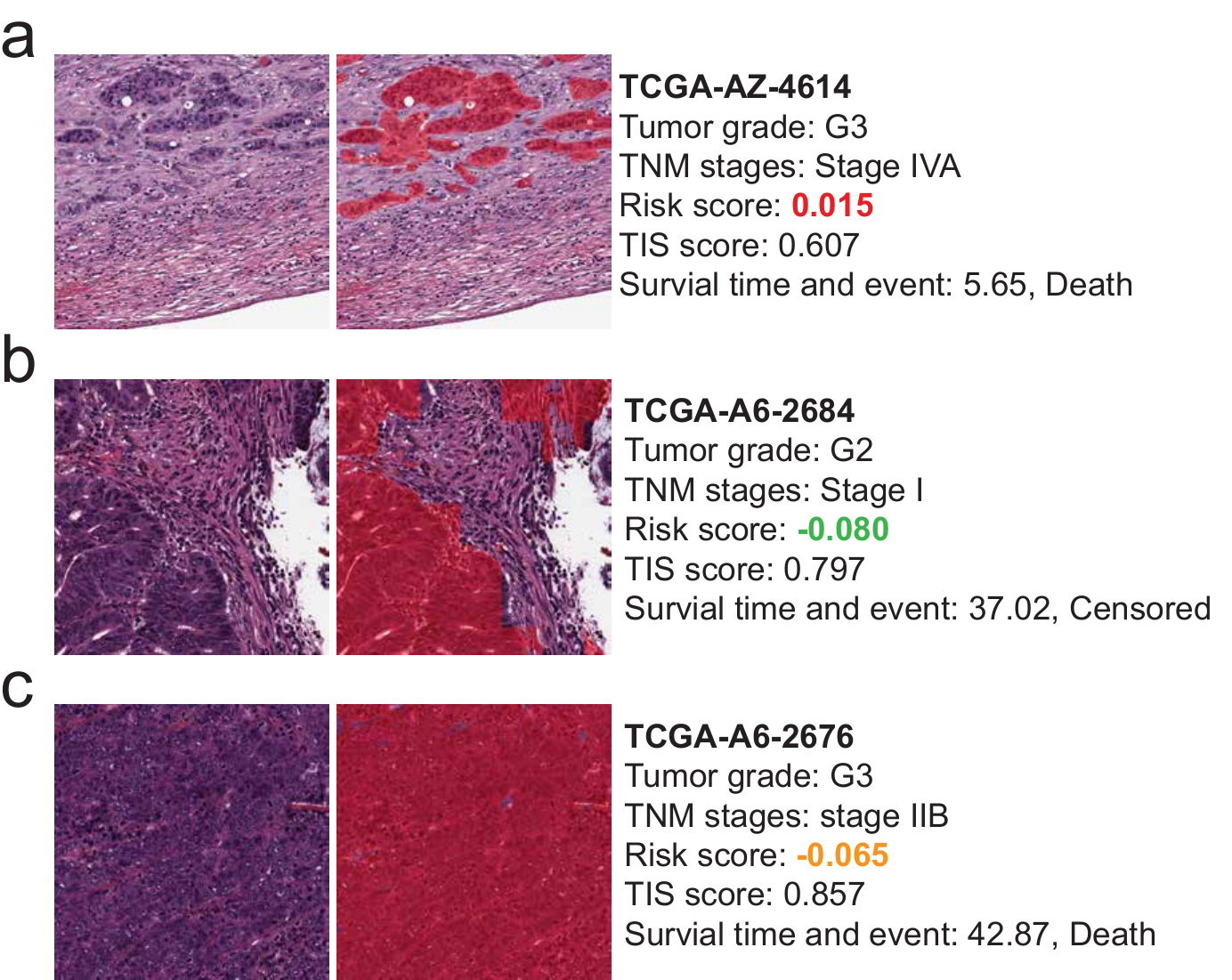}
\caption{\textbf{Extended Data Figure~4. Three representative H\&E patches and corresponding tumor segmentation overlays from TCGA colorectal cancer cases illustrating low, intermediate, and high Tumor Interation Score (TIS).} Risk score represents a relative risk ranking derived from our survival model and is independent of sign; higher risk values indicate worse survival outcomes. TNM denotes Tumor--Node--Metastasis staging.
\textbf{a}, The low TIS case is characterized by pronounced tumor budding and infiltrative growth at the invasive front, extending toward the serosal surface, consistent with poorly differentiated tumor and elevated clinical risk.
\textbf{b}, The intermediate TIS case exhibits a well-demarcated tumor boundary with prominent lymphocytic infiltration in the surrounding stroma, reflecting spatially confined tumor growth and the lowest predicted risk among the examples.
\textbf{c}, The high TIS case corresponds to a large, highly cohesive tumor mass with extensive tumor area, where excessive tumor burden is associated with a renewed increase in predicted risk.}
\label{sup:fig4}
\end{figure}

\begin{table}[tbp]
\captionsetup{labelformat=empty}
\begin{center}
\begin{tabular}{ c >{\raggedright\arraybackslash}p{12cm}}\hline

 Data Source & Link \\\hline
AGGC~\cite{huo2022comprehensive} & \url{https://medicalimaging.ai/challenges/aggc22} \\
BACH~\cite{polonia2019bach} & \url{https://zenodo.org/records/3632035}  \\
BCNB~\cite{xu2021predicting} & \url{https://bcnb.grand-challenge.org/} \\
BRCA~\cite{amgad2019structured} & \url{https://github.com/PathologyDataScience/BCSS} \\
CoCaHis~\cite{sitnik2021dataset} & \url{https://cocahis.irb.hr/} \\
DROID~\cite{stadler2021proactive} & \url{https://datahub.aida.scilifelab.se/10.23698/aida/drsk} \\
DROV~\cite{stadler2021proactive} & \url{https://datahub.aida.scilifelab.se/10.23698/aida/drov} \\
DRYAD~\cite{cruz2018high} & \url{https://datadryad.org/dataset/doi:10.5061/dryad.1g2nt41#citations}  \\
DigestPath~\cite{da2022digestpath,li2019signet} & \url{https://digestpath2019.grand-challenge.org/} \\
GlaS~\cite{sirinukunwattana2017gland} & \url{https://www.kaggle.com/datasets/sani84/glasmiccai2015-gland-segmentation} \\
Histo-Seg~\cite{abdul2024histo} & \url{https://data.mendeley.com/datasets/vccj8mp2cg/1} \\
HuBMAP~\cite{howardhubmap} & 
\url{https://www.kaggle.com/competitions/hubmap-hacking-the-human-vasculature/data} \\
KPMP~\cite{KPMP} & \url{https://www.kpmp.org/available-data} \\
KPIs~\cite{deng2025kpis} & \url{https://www.synapse.org/Synapse:syn54077668/wiki/626475} \\
Lung\_seg~\cite{kludt2024next} &  \url{https://zenodo.org/records/12818382} \\
ORCA~\cite{martino2020deep} &  \url{https://sites.google.com/unibas.it/orca}  \\
PANDA~\cite{bulten2022artificial} & \url{https://www.kaggle.com/competitions/prostate-cancer-grade-assessment/overview/citation} \\
TIGER\_wsibulk~\cite{van2025tumor} & \url{https://tiger.grand-challenge.org/Data/} \\
TIGER\_wsirois~\cite{van2025tumor} & \url{https://tiger.grand-challenge.org/Data/} \\
WSSS4LUAD~\cite{Han2022Multilayer} & \url{https://wsss4luad.grand-challenge.org/} \\
prostate\_gland~\cite{oner2022ai} & \url{https://zenodo.org/records/5971764} \\ \hline

\end{tabular}
\end{center}
\caption{\textbf{Extended Data Table~1. Overview of internal datasets used in our VISTA-PATH Data for pathology image segmentation.}}
\label{sup:internal_data_source}
\end{table}

\begin{table}[tbp]
\captionsetup{labelformat=empty}
\begin{center}
\begin{tabular}{ c c  }\hline

Data Source  & Link \\\hline
 \multicolumn{2}{l}{\textit{\textcolor{gray}{Public Pathology Datasets}}} \\
LungHP~\cite{li2020deep} &  \url{https://acdc-lunghp.grand-challenge.org/}  \\
OCDC~\cite{dos2023hematoxylin} & \url{https://github.com/dalifreire/tumor_regions_segmentation} \\ \hline
\multicolumn{2}{l}{\textit{\textcolor{gray}{Visium HD Datasets}}} \\
Visium\_Colon~\cite{oliveira2025high} &  \url{https://www.10xgenomics.com/datasets} \\
Visium\_Kidney~\cite{oliveira2025high} &  \url{https://www.10xgenomics.com/datasets} \\
Visium\_Lung~\cite{oliveira2025high} & \url{https://www.10xgenomics.com/datasets}  \\
Visium\_Pancreas~\cite{oliveira2025high} &  \url{https://www.10xgenomics.com/datasets}  \\
Visium\_Prostate~\cite{oliveira2025high} &  \url{https://www.10xgenomics.com/datasets} \\ \hline
\multicolumn{2}{l}{\textit{\textcolor{gray}{Xenium Datasets}}} \\
Brain-Immuno~\cite{janesick2023high} & \url{https://www.10xgenomics.com/datasets} \\
Breast-5K~\cite{janesick2023high} & \url{https://www.10xgenomics.com/datasets} \\
Breast-CA~\cite{janesick2023high} & \url{https://www.10xgenomics.com/datasets} \\
Breast-Entire~\cite{janesick2023high} &  \url{https://www.10xgenomics.com/datasets}\\
Cervical-5kPathway~\cite{janesick2023high} & \url{https://www.10xgenomics.com/datasets} \\
Colon\_Preview~\cite{janesick2023high} & \url{https://www.10xgenomics.com/datasets} \\
Heart-MultiCancer~\cite{janesick2023high} & \url{https://www.10xgenomics.com/datasets} \\
Kidney-MultiCancer~\cite{janesick2023high} & \url{https://www.10xgenomics.com/datasets} \\
Liver-Healthy~\cite{janesick2023high} & \url{https://www.10xgenomics.com/datasets} \\
Lung-Cancer~\cite{janesick2023high} & \url{https://www.10xgenomics.com/datasets} \\
Lung-GExPanel~\cite{janesick2023high} & \url{https://www.10xgenomics.com/datasets} \\
Lung-Immuno~\cite{janesick2023high} & \url{https://www.10xgenomics.com/datasets} \\
Ovarian-5KPathway~\cite{janesick2023high} & \url{https://www.10xgenomics.com/datasets} \\
Ovarian-Immuno~\cite{janesick2023high} & \url{https://www.10xgenomics.com/datasets} \\
Pancreatic-Immuno~\cite{janesick2023high} & \url{https://www.10xgenomics.com/datasets} \\
Pancreatic-Cancer~\cite{janesick2023high} & \url{https://www.10xgenomics.com/datasets} \\
Pancreatic-Modal~\cite{janesick2023high} &  \url{https://www.10xgenomics.com/datasets}\\
Skin-5kPathway~\cite{janesick2023high} & \url{https://www.10xgenomics.com/datasets} \\
Skin-GExPanel~\cite{janesick2023high} & \url{https://www.10xgenomics.com/datasets} \\
Skin-MultiCancer~\cite{janesick2023high} & \url{https://www.10xgenomics.com/datasets} \\
Tonsil-Follicular~\cite{janesick2023high} &  \url{https://www.10xgenomics.com/datasets}\\
Tonsil-Reactive~\cite{janesick2023high} & \url{https://www.10xgenomics.com/datasets} \\
Xenium-1k~\cite{janesick2023high} & \url{https://www.10xgenomics.com/datasets} \\ \hline

\end{tabular}
\end{center}
\caption{\textbf{Extended Data Table~2. Overview of external datasets for external segmentation evaluation.}}
\label{sup:external_data_source}
\end{table}

\begin{table}[tbp]
\captionsetup{labelformat=empty}
\begin{center}
\begin{tabular}{  c c c c c c }\hline

 & VISTA-PATH & VISTA-PATH (w/o box) & MedSAM & BiomedParse & Res2Net \\\cline{2-6}
  & Dice $(\%)$ $\uparrow$ & Dice $(\%)$ $\uparrow$ & Dice $(\%)$ $\uparrow$ & Dice $(\%)$ $\uparrow$ & Dice $(\%)$ $\uparrow$ \\ \hline
AGGC & \textbf{0.845} & 0.580 & 0.837 & 0.264 & 0.497 \\
BACH & 0.893 & 0.739 & \textbf{0.902} & 0.290 & 0.308 \\
BCNB & \textbf{0.874} & 0.729 & 0.822 & 0.525 & 0.606 \\
BRCA & \textbf{0.690} & 0.592 & 0.355 & 0.022 & 0.347 \\
CoCaHis & 0.835 & 0.831 & 0.349 & 0.695 & \textbf{0.854} \\
DROID & \textbf{0.767} & 0.670 & 0.723 & 0.174 & 0.519 \\
DROV & \textbf{0.878} & 0.796 & 0.851 & 0.489 & 0.124 \\
DRYAD & \textbf{0.872} & 0.823 & 0.866 & 0.467 & 0.700 \\
DigestPath & \textbf{0.888} & 0.808 & 0.750 & 0.722 & 0.728 \\
GlaS & 0.886 & 0.884 & 0.522 & \textbf{0.893} & 0.735 \\
Histo-Seg & \textbf{0.508} & 0.441 & 0.397 & 0.218 & 0.462 \\
HuBMAP & \textbf{0.582} & 0.549 & 0.421 & 0.398 & 0.448 \\
KPMP & \textbf{0.779} & 0.731 & 0.377 & 0.145 & 0.496 \\
KPIs & \textbf{0.952} & 0.924 & 0.790 & 0.382 & 0.697 \\
Lung\_seg & \textbf{0.663} & 0.483 & 0.249 & 0.194 & 0.367 \\
ORCA & 0.602 & \textbf{0.608} & 0.338 & 0.243 & 0.224 \\
PANDA & \textbf{0.930} & 0.913 & 0.772 & 0.417 & 0.849 \\
TIGER\_wsibulk & \textbf{0.892} & 0.846 & 0.863 & 0.303 & 0.432 \\
TIGER\_wsirois & \textbf{0.590} & 0.433 & 0.417 & 0.079 & 0.208 \\
WSSS4LUAD & 0.453 & 0.432 & 0.180 & 0.304 & \textbf{0.541} \\
prostate\_gland\_segmentation & \textbf{0.842} & 0.841 & 0.425 & 0.727 & 0.790 \\
\hline
\textbf{Average} 
& \textbf{0.772} 
& 0.698 
& 0.581 
& 0.379 
& 0.521 \\\hline

\end{tabular}
\end{center}
\caption{\textbf{Extended Data Table~3. Segmentation comparison results on held-out internal datasets.}}
\label{sup:table_internal}
\end{table}

\begin{table}[htbp]
\centering
\captionsetup{labelformat=empty}
\begin{tabular}{l|ccc|cc}
\hline
Organ & VISTA-PATH & MedSAM & BiomedParse & $\Delta$ MedSAM & $\Delta$ BiomedParse \\
\hline
Breast & \textbf{0.802} & 0.704 & 0.281 & +0.098 & +0.521 \\
Colon & \textbf{0.887} & 0.636 & 0.808 & +0.251 & +0.080 \\
Kidney & \textbf{0.771} & 0.529 & 0.308 & +0.242 & +0.463 \\
Liver & \textbf{0.835} & 0.349 & 0.695 & +0.486 & +0.140 \\
Lung & \textbf{0.558} & 0.215 & 0.249 & +0.344 & +0.309 \\
Oral & \textbf{0.602} & 0.338 & 0.243 & +0.264 & +0.359 \\
Ovary & \textbf{0.878} & 0.851 & 0.489 & +0.027 & +0.389 \\
Prostate & \textbf{0.872} & 0.678 & 0.469 & +0.194 & +0.403 \\
Skin & \textbf{0.637} & 0.560 & 0.196 & +0.077 & +0.441 \\
\hline
\end{tabular}
\caption{\textbf{Extended Data Table~4. Dice scores across organs on the Internal Evaluation dataset.} $\Delta$ MedSAM = VISTA-PATH $-$ MedSAM; $\Delta$ BiomedParse = VISTA-PATH $-$ BiomedParse. Positive values indicate VISTA-PATH outperforms the compared method.}
\label{sup:organ_internal}
\end{table}

\begin{table}[htbp]
\centering
\captionsetup{labelformat=empty}
\begin{tabular}{ll|ccc|cc}
\hline
Category & Organ & VISTA-PATH & MedSAM & BiomedParse & $\Delta$ MedSAM & $\Delta$ BiomedParse \\
\hline
\multirow{7}{*}{Tumor Related} & Breast & \textbf{0.862} & 0.828 & 0.196 & +0.034$^{***}$ & +0.666$^{***}$ \\
 & Colon & \textbf{0.864} & 0.754 & 0.634 & +0.109$^{***}$ & +0.230$^{***}$ \\
 & Liver & \textbf{0.815} & 0.563 & 0.623 & +0.251$^{***}$ & +0.191$^{***}$ \\
 & Lung & \textbf{0.553} & 0.215 & 0.256 & +0.338$^{***}$ & +0.297$^{***}$ \\
 & Oral & \textbf{0.659} & 0.362 & 0.254 & +0.297$^{***}$ & +0.405$^{***}$ \\
 & Ovary & \textbf{0.905} & 0.878 & 0.495 & +0.027$^{***}$ & +0.410$^{***}$ \\
 & Prostate & \textbf{0.838} & 0.652 & 0.325 & +0.186$^{***}$ & +0.512$^{***}$ \\
\hline
\multirow{5}{*}{Microenvironment Related} & Breast & \textbf{0.575} & 0.462 & 0.054 & +0.113$^{***}$ & +0.521$^{***}$ \\
 & Kidney & \textbf{0.692} & 0.625 & 0.078 & +0.067$^{***}$ & +0.614$^{***}$ \\
 & Lung & \textbf{0.548} & 0.130 & 0.079 & +0.418$^{***}$ & +0.469$^{***}$ \\
 & Prostate & 0.852 & \textbf{0.883} & 0.041 & -0.031 & +0.811$^{***}$ \\
 & Skin & \textbf{0.665} & 0.648 & 0.036 & +0.018 & +0.630$^{***}$ \\
\hline
\multirow{8}{*}{Normal Anatomical Related} & Breast & \textbf{0.701} & 0.388 & 0.063 & +0.313$^{***}$ & +0.637$^{***}$ \\
 & Colon & 0.878 & 0.351 & \textbf{0.882} & +0.527$^{***}$ & -0.004 \\
 & Kidney & \textbf{0.787} & 0.383 & 0.091 & +0.405$^{***}$ & +0.697$^{***}$ \\
 & Lung & \textbf{0.565} & 0.144 & 0.063 & +0.421$^{***}$ & +0.503$^{***}$ \\
 & Oral & \textbf{0.540} & 0.146 & 0.102 & +0.394$^{***}$ & +0.438$^{***}$ \\
 & Ovary & 0.816 & \textbf{0.861} & 0.027 & -0.045 & +0.789$^{***}$ \\
 & Prostate & \textbf{0.926} & 0.735 & 0.155 & +0.190$^{***}$ & +0.771$^{***}$ \\
 & Skin & \textbf{0.826} & 0.793 & 0.149 & +0.033$^{***}$ & +0.677$^{***}$ \\
\hline
\end{tabular}
\caption{\textbf{Extended Data Table~5. Dice scores across organs, grouped into three major tissue categories, on the internal evaluation dataset.} $\Delta$ MedSAM = VISTA-PATH $-$ MedSAM; $\Delta$ BiomedParse = VISTA-PATH $-$ BiomedParse. Statistical significance is assessed using two-sided Student's $t$-test. $^{*}P < 0.05$; $^{**}P < 1 \times 10^{-2}$; $^{***}P < 1 \times 10^{-3}$.}
\label{sup:organ_tissue_internal}
\end{table}

\begin{table}[tbp]
\captionsetup{labelformat=empty}
\begin{center}
\begin{tabular}{  c c c c c }\hline

 & VISTA-PATH & VISTA-PATH (w/o box) & MedSAM & BiomedParse  \\\cline{2-5}
  & Dice $(\%)$ $\uparrow$ & Dice $(\%)$ $\uparrow$ & Dice $(\%)$ $\uparrow$ & Dice $(\%)$ $\uparrow$   \\ \hline
  \multicolumn{4}{l}{\textit{\textcolor{gray}{Public Pathology Datasets}}} \\
LungHP & \textbf{0.495} & 0.346 & 0.121 & 0.129   \\
OCDC & \textbf{0.802} & 0.787 & 0.528 & 0.751 \\ \hline
\multicolumn{4}{l}{\textit{\textcolor{gray}{Visium HD Datasets}}} \\
Visium\_Colon & \textbf{0.694} & 0.621 & 0.556 & 0.242 \\
Visium\_Kidney &  \textbf{0.832} & 0.579 & 0.812 &	0.266   \\
Visium\_Lung & \textbf{0.506} & 0.422 & 0.324 & 0.149    \\
Visium\_Pancreas & \textbf{0.373} & 0.253 & 0.329 & 0.093    \\
Visium\_Prostate & \textbf{0.249} & 0.143 & 0.217 & 0.064   \\ \hline
\multicolumn{4}{l}{\textit{\textcolor{gray}{Xenium Datasets}}} \\
Brain-Immuno & \textbf{0.231} & 0.193 & 0.154 & 0.150 \\
Breast-5K & \textbf{0.313} & 0.246 & 0.202 & 0.245 \\
Breast-CA & \textbf{0.388} & 0.230 & 0.314 & 0.112 \\
Breast-Entire & \textbf{0.374} & 0.228 & 0.296 & 0.082 \\
Cervical-5kPathway & \textbf{0.362} & 0.286 & 0.290 & 0.275 \\
Colon\_Preview & \textbf{0.610} & 0.510 & 0.537 & 0.209 \\
Heart-MultiCancer & \textbf{0.410} & 0.288 & 0.337 & 0.131 \\
Kidney-MultiCancer & \textbf{0.427} & 0.267 & 0.363 & 0.104 \\
Liver-Healthy & \textbf{0.346} & 0.121 & 0.261 & 0.052 \\
Lung-Cancer & \textbf{0.369} & 0.251 & 0.298 & 0.183 \\
Lung-GExPanel & \textbf{0.392} & 0.302 & 0.304 & 0.247 \\
Lung-Immuno & \textbf{0.475} & 0.337 & 0.372 & 0.224 \\
Ovarian-5KPathway & \textbf{0.301} & 0.251 & 0.244 & 0.268 \\
Ovarian-Immuno & \textbf{0.404} & 0.312 & 0.386 & 0.175 \\
Pancreatic-Immuno & \textbf{0.527} & 0.446 & 0.487 & 0.251 \\
Pancreatic-Cancer & \textbf{0.555} & 0.378 & 0.514 & 0.179 \\
Pancreatic-Modal & \textbf{0.449} & 0.355 & 0.371 & 0.306 \\
Skin-5kPathway & \textbf{0.408} & 0.297 & 0.361 & 0.289 \\
Skin-GExPanel & \textbf{0.620} & 0.475 & 0.465 & 0.159 \\
Skin-MultiCancer & \textbf{0.454} & 0.351 & 0.408 & 0.333 \\
Tonsil-Follicular & \textbf{0.599} & 0.461 & 0.546 & 0.184 \\
Tonsil-Reactive & \textbf{0.474} & 0.291 & 0.435 & 0.172 \\
Xenium-1k & \textbf{0.523} & 0.417 & 0.471 & 0.186 \\
\hline 
\textbf{Average}
& \textbf{0.454}
& 0.340
& 0.373
& 0.199 \\ \hline

\end{tabular}
\end{center}
\caption{\textbf{Extended Data Table~6. Segmentation comparison results on external datasets.}}
\label{sup:table_external}
\end{table}

\begin{table}[htbp]
\captionsetup{labelformat=empty}
\centering
\begin{tabular}{l|ccc|cc}
\hline
Organ & VISTA-PATH & MedSAM & BiomedParse & $\Delta$ MedSAM & $\Delta$ BiomedParse \\
\hline
Brain & \textbf{0.231} & 0.154 & 0.150 & +0.077 & +0.081 \\
Breast & \textbf{0.479} & 0.429 & 0.182 & +0.050 & +0.297 \\
Cervical & \textbf{0.362} & 0.290 & 0.275 & +0.071 & +0.087 \\
Colon & \textbf{0.652} & 0.546 & 0.226 & +0.106 & +0.426 \\
Heart & \textbf{0.410} & 0.337 & 0.131 & +0.073 & +0.279 \\
Kidney & \textbf{0.630} & 0.587 & 0.185 & +0.042 & +0.445 \\
Lung & \textbf{0.447} & 0.284 & 0.186 & +0.163 & +0.261 \\
Oral & \textbf{0.802} & 0.528 & 0.751 & +0.275 & +0.051 \\
Ovarian & \textbf{0.353} & 0.315 & 0.222 & +0.037 & +0.131 \\
Pancreas & \textbf{0.476} & 0.425 & 0.207 & +0.050 & +0.268 \\
Prostate & \textbf{0.249} & 0.217 & 0.064 & +0.033 & +0.185 \\
Skin & \textbf{0.494} & 0.411 & 0.260 & +0.083 & +0.234 \\
Tonsil & \textbf{0.537} & 0.490 & 0.178 & +0.046 & +0.358 \\
\hline
\end{tabular}
\caption{\textbf{Extended Data Table~7. Dice scores across organs on the external evaluation datasets.} $\Delta$ MedSAM = VISTA-PATH $-$ MedSAM; $\Delta$ BiomedParse = VISTA-PATH $-$ BiomedParse. Positive values indicate VISTA-PATH outperforms the compared method.}
\label{sup:organ_external}
\end{table}

\begin{table}[htbp]
\captionsetup{labelformat=empty}
\centering
\begin{tabular}{ll|ccc|cc}
\hline
Category & Organ & VISTA-PATH & MedSAM & BiomedParse & $\Delta$ MedSAM & $\Delta$ BiomedParse \\
\hline
\multirow{9}{*}{Tumor Related} & Brain & \textbf{0.217} & 0.125 & 0.203 & +0.092$^{***}$ & +0.014$^{***}$ \\
 & Breast & \textbf{0.511} & 0.383 & 0.150 & +0.128$^{***}$ & +0.361$^{***}$ \\
 & Cervical & \textbf{0.787} & 0.518 & 0.364 & +0.269$^{***}$ & +0.423$^{***}$ \\
 & Colon & \textbf{0.709} & 0.662 & 0.430 & +0.048$^{***}$ & +0.279$^{***}$ \\
 & Lung & \textbf{0.544} & 0.401 & 0.323 & +0.143$^{***}$ & +0.221$^{***}$ \\
 & Oral & \textbf{0.720} & 0.251 & 0.595 & +0.469$^{***}$ & +0.125$^{**}$ \\
 & Ovary & \textbf{0.703} & 0.446 & 0.411 & +0.257$^{***}$ & +0.292$^{***}$ \\
 & Pancreas & \textbf{0.622} & 0.604 & 0.217 & +0.018 & +0.405$^{***}$ \\
 & Prostate & \textbf{0.465} & 0.371 & 0.240 & +0.095$^{***}$ & +0.225$^{***}$ \\
\hline
\multirow{11}{*}{Microenvironment Related} & Breast & \textbf{0.304} & 0.265 & 0.074 & +0.040$^{***}$ & +0.230$^{***}$ \\
 & Cervical & \textbf{0.309} & 0.262 & 0.127 & +0.046$^{***}$ & +0.181$^{***}$ \\
 & Colon & \textbf{0.654} & 0.517 & 0.046 & +0.137$^{***}$ & +0.608$^{***}$ \\
 & Heart & \textbf{0.277} & 0.252 & 0.007 & +0.024$^{**}$ & +0.270$^{***}$ \\
 & Kidney & \textbf{0.299} & 0.298 & 0.038 & +0.001 & +0.261$^{***}$ \\
 & Lung & \textbf{0.346} & 0.259 & 0.060 & +0.087$^{***}$ & +0.286$^{***}$ \\
 & Ovary & \textbf{0.246} & 0.216 & 0.075 & +0.029$^{***}$ & +0.171$^{***}$ \\
 & Pancreas & \textbf{0.439} & 0.383 & 0.085 & +0.056$^{***}$ & +0.353$^{***}$ \\
 & Prostate & \textbf{0.235} & 0.156 & 0.039 & +0.079$^{***}$ & +0.196$^{***}$ \\
 & Skin & \textbf{0.457} & 0.335 & 0.093 & +0.122$^{***}$ & +0.364$^{***}$ \\
 & Tonsil & \textbf{0.550} & 0.499 & 0.235 & +0.051$^{***}$ & +0.315$^{***}$ \\
\hline
\multirow{12}{*}{Normal Anatomical Related} & Brain & \textbf{0.187} & 0.108 & 0.128 & +0.078$^{***}$ & +0.058$^{***}$ \\
 & Breast & \textbf{0.553} & 0.453 & 0.326 & +0.100$^{***}$ & +0.227$^{***}$ \\
 & Cervical & \textbf{0.551} & 0.533 & 0.514 & +0.019 & +0.037 \\
 & Colon & \textbf{0.694} & 0.628 & 0.126 & +0.066$^{***}$ & +0.568$^{***}$ \\
 & Heart & \textbf{0.454} & 0.353 & 0.023 & +0.101$^{***}$ & +0.431$^{***}$ \\
 & Kidney & \textbf{0.562} & 0.483 & 0.063 & +0.079$^{***}$ & +0.499$^{***}$ \\
 & Lung & \textbf{0.350} & 0.299 & 0.141 & +0.051$^{***}$ & +0.209$^{***}$ \\
 & Ovary & 0.169 & \textbf{0.191} & 0.121 & -0.023$^{***}$ & +0.048$^{***}$ \\
 & Pancreas & \textbf{0.578} & 0.516 & 0.286 & +0.061$^{***}$ & +0.292$^{***}$ \\
 & Prostate & 0.112 & \textbf{0.162} & 0.099 & -0.050$^{***}$ & +0.013$^{**}$ \\
 & Skin & \textbf{0.490} & 0.488 & 0.081 & +0.003 & +0.409$^{***}$ \\
 & Tonsil & \textbf{0.506} & 0.487 & 0.223 & +0.019$^{**}$ & +0.283$^{***}$ \\
\hline
\end{tabular}
\caption{\textbf{Extended Data Table~8. Dice scores across organs, grouped into three major tissue categories, on the external evaluation dataset.} $\Delta$ MedSAM = VISTA-PATH $-$ MedSAM; $\Delta$ BiomedParse = VISTA-PATH $-$ BiomedParse. Statistical significance is assessed using two-sided Student's $t$-test. $^{*}P < 0.05$; $^{**}P < 1 \times 10^{-2}$; $^{***}P < 1 \times 10^{-3}$.}
\label{sup:organ_tissue_external}
\end{table}

\begin{table}[htbp]
\captionsetup{labelformat=empty}
\centering
\begin{tabular}{ll|ccccccc}
\hline
Dataset & Method & \multicolumn{7}{c}{Number of Patches} \\
\hline
\multirow{4}{*}{Visium\_Colon} &  & 0 & 226 & 321 & 1051 & 1079 & 1175 &  \\
 & VISTA-PATH & \textbf{0.629} & \textbf{0.627} & \textbf{0.665} & \textbf{0.861} & \textbf{0.857} & \textbf{0.867} &  \\
 & Patch & 0.568 & 0.559 & 0.595 & 0.804 & 0.797 & 0.808 &  \\
 & MedSAM & 0.563 & 0.561 & 0.584 & 0.765 & 0.755 & 0.769 &  \\
\hline
\multirow{4}{*}{Visium\_Kidney} &  & 0 & 10 & 33 & 65 & 146 & 158 & 172 \\
 & VISTA-PATH & \textbf{0.118} & 0.117 & 0.198 & \textbf{0.343} & 0.401 & \textbf{0.587} & \textbf{0.563} \\
 & Patch & 0.118 & 0.112 & \textbf{0.228} & 0.308 & \textbf{0.418} & 0.533 & 0.534 \\
 & MedSAM & 0.118 & \textbf{0.118} & 0.159 & 0.270 & 0.291 & 0.476 & 0.462 \\
\hline
\multirow{4}{*}{Visium\_Prostate} &  & 0 & 81 & 136 & 240 & 288 & 296 &  \\
 & VISTA-PATH & \textbf{0.424} & \textbf{0.424} & \textbf{0.603} & \textbf{0.805} & \textbf{0.818} & \textbf{0.810} &  \\
 & Patch & 0.409 & 0.408 & 0.576 & 0.769 & 0.776 & 0.763 &  \\
 & MedSAM & 0.410 & 0.410 & 0.527 & 0.774 & 0.788 & 0.775 &  \\
\hline
\end{tabular}
\caption{\textbf{Extended Data Table~9. Human-in-the-loop results on Visium HD datasets.} Dice scores are reported. }
\label{tab:dice_visium}
\end{table}

\begin{table}[htbp]
\captionsetup{labelformat=empty}
\centering
\begin{tabular}{ll | cccccccccc}
\hline
Dataset & Method & \multicolumn{10}{c}{Number of Patches} \\
\hline
\multirow{4}{*}{Cervical-5kPathway} &  & 0 & 153 & 932 & 1332 & 1350 &  &  &  &  &  \\
 & VISTA-PATH & \textbf{0.429} & \textbf{0.429} & \textbf{0.773} & \textbf{0.758} & \textbf{0.771} &  &  &  &  &  \\
 & Patch & 0.372 & 0.372 & 0.667 & 0.650 & 0.724 &  &  &  &  &  \\
 & MedSAM & 0.364 & 0.364 & 0.650 & 0.670 & 0.586 &  &  &  &  &  \\
\hline
\multirow{4}{*}{Ovarian-5KPathway} &  & 0 & 71 & 313 & 376 & 386 & 390 &  &  &  &  \\
 & VISTA-PATH & \textbf{0.375} & \textbf{0.366} & \textbf{0.851} & \textbf{0.845} & \textbf{0.838} & \textbf{0.843} &  &  &  &  \\
 & Patch & 0.352 & 0.335 & 0.773 & 0.745 & 0.734 & 0.748 &  &  &  &  \\
 & MedSAM & 0.345 & 0.342 & 0.725 & 0.725 & 0.747 & 0.750 &  &  &  &  \\
\hline
\multirow{4}{*}{Breast-5K} &  & 0 & 166 & 543 & 614 & 744 & 789 & 877 &  &  &  \\
 & VISTA-PATH & \textbf{0.265} & \textbf{0.265} & \textbf{0.556} & \textbf{0.584} & \textbf{0.558} & \textbf{0.641} & \textbf{0.693} &  &  &  \\
 & Patch & 0.242 & 0.242 & 0.533 & 0.560 & 0.457 & 0.558 & 0.610 &  &  &  \\
 & MedSAM & 0.232 & 0.232 & 0.421 & 0.435 & 0.473 & 0.508 & 0.578 &  &  &  \\
\hline
\multirow{4}{*}{Skin-MultiCancer} &  & 0 & 199 & 537 & 702 & 767 &  &  &  &  &  \\
 & VISTA-PATH & \textbf{0.418} & 0.267 & 0.269 & 0.474 & \textbf{0.583} &  &  &  &  &  \\
 & Patch & 0.339 & \textbf{0.268} & \textbf{0.277} & \textbf{0.507} & 0.537 &  &  &  &  &  \\
 & MedSAM & 0.393 & 0.255 & 0.261 & 0.427 & 0.533 &  &  &  &  &  \\
\hline
\multirow{4}{*}{Lung-Immuno} &  & 0 & 55 & 74 & 94 & 139 & 239 & 274 & 354 & 555 &  \\
 & VISTA-PATH & \textbf{0.413} & \textbf{0.413} & \textbf{0.434} & \textbf{0.616} & \textbf{0.666} & \textbf{0.709} & \textbf{0.703} & \textbf{0.723} & \textbf{0.745} &  \\
 & Patch & 0.335 & 0.335 & 0.375 & 0.543 & 0.569 & 0.646 & 0.672 & 0.678 & 0.716 &  \\
 & MedSAM & 0.319 & 0.319 & 0.328 & 0.417 & 0.468 & 0.585 & 0.547 & 0.590 & 0.626 &  \\
\hline
\multirow{4}{*}{Lung-GExPanel} &  & 0 & 296 & 345 & 1027 & 1081 & 1285 & 1467 & 1599 & 1707 & 1815 \\
 & VISTA-PATH & \textbf{0.421} & \textbf{0.421} & \textbf{0.501} & \textbf{0.467} & \textbf{0.495} & 0.539 & \textbf{0.605} & \textbf{0.607} & \textbf{0.553} & \textbf{0.574} \\
 & Patch & 0.377 & 0.376 & 0.473 & 0.447 & 0.469 & \textbf{0.542} & 0.595 & 0.606 & 0.541 & 0.559 \\
 & MedSAM & 0.373 & 0.373 & 0.415 & 0.396 & 0.415 & 0.474 & 0.539 & 0.555 & 0.496 & 0.504 \\
\hline
\end{tabular}
\caption{\textbf{Extended Data Table~10. Human-in-the-loop results on Xenium datasets.} Dice scores are reported.}
\label{tab:dice_xenium}
\end{table}

\end{document}